\documentclass{article}

\PassOptionsToPackage{numbers, compress}{natbib}

\usepackage[final]{neurips_2022}

\usepackage[utf8]{inputenc} %
\usepackage[T1]{fontenc}    %
\usepackage[hypertexnames=false,colorlinks=true,citecolor=NavyBlue]{hyperref}       %
\usepackage{url}            %
\usepackage{booktabs}       %
\usepackage{amsfonts}       %
\usepackage{nicefrac}       %
\usepackage{microtype}      %
\usepackage[table,xcdraw,dvipsnames]{xcolor}         %
\usepackage{graphicx}
\usepackage{subcaption}
\usepackage{wrapfig}
\usepackage{multirow}
\usepackage{makecell}
\usepackage{pifont}
\usepackage{wrapfig,lipsum,booktabs}
\newcommand{\cmark}{\ding{51}}%
\newcommand{\xmark}{\ding{55}}%
\NewDocumentCommand{\anote}{}{\makebox[0pt][l]{$^*$}}

\newcommand{\model}{\textit{FrozenBiLM}}

\definecolor{greencode}{RGB}{13, 144, 79}

\title{Zero-Shot Video Question Answering via \\ 
Frozen Bidirectional Language Models}

\author{Antoine Yang$^{1,2}$, Antoine Miech$^{3}$, Josef Sivic$^{4}$, Ivan Laptev$^{1,2}$, Cordelia Schmid$^{1,2}$
\\ 
$^1$Inria Paris \quad $^2$D\'{e}partement d'informatique de l'ENS, CNRS, PSL Research University
\\ 
\quad $^3$DeepMind \quad $^4$CIIRC CTU Prague
\\
\url{https://antoyang.github.io/frozenbilm.html}
\vspace{-0.5cm}
}

\begin{document}

\maketitle

\begin{abstract}
Video question answering (VideoQA) is a complex task that requires diverse multi-modal data for training.
Manual annotation of question and answers for videos, however, is tedious and prohibits scalability.
To tackle this problem, recent methods consider zero-shot settings with no manual annotation of visual question-answer.
In particular, a promising approach adapts \emph{frozen autoregressive} language models pretrained on Web-scale text-only data to multi-modal inputs.
In contrast, we here build on \emph{frozen bidirectional} language models (BiLM) and show that such an approach provides a stronger and cheaper alternative for zero-shot VideoQA.
In particular, (i)~we combine visual inputs with the frozen BiLM using light trainable modules, (ii)~we train such modules using Web-scraped multi-modal data, and finally
(iii)~we perform zero-shot VideoQA inference through masked language modeling, where the masked text is the answer to a given question.
Our proposed approach, \model{}, outperforms the state of the art in zero-shot VideoQA by a significant margin on a variety of datasets, including LSMDC-FiB, iVQA, MSRVTT-QA, MSVD-QA, ActivityNet-QA, TGIF-FrameQA, How2QA and TVQA.
It also demonstrates competitive performance in the few-shot and fully-supervised setting.
Our code and models are publicly available at~\cite{frozenbilmwebpage}.
\end{abstract}

\footnotetext[4]{Czech Institute of Informatics, Robotics and Cybernetics at the Czech Technical University in Prague.}

\begin{figure*}[h]
\centering
\vspace{-0.6cm}
\includegraphics[width=1.\linewidth]{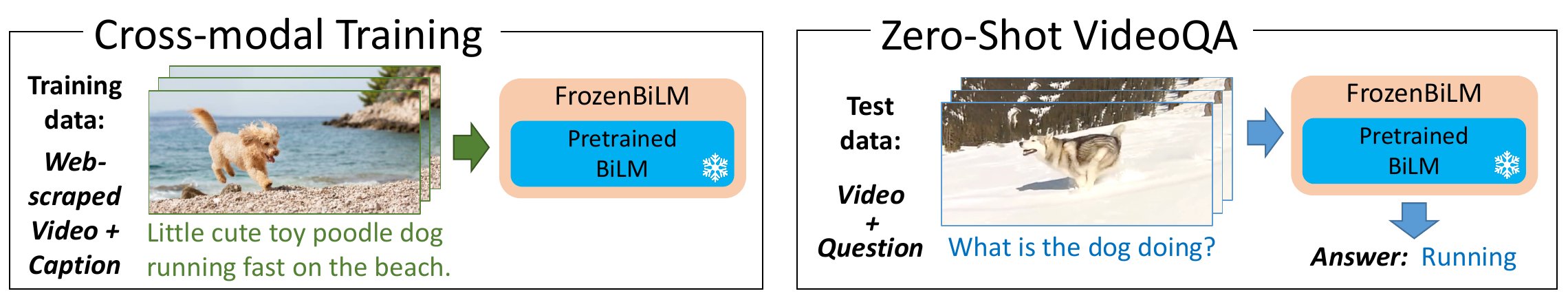}
\vspace{-0.5cm}
\caption{
Our model \model{} builds on a pretrained and \emph{frozen} bidirectional language model (BiLM), and is trained from Web-scraped video-caption pairs. \model{} excels in the zero-shot video question answering task without using any explicit visual question-answer supervision. 
}
\vspace{-0.3cm}
\label{fig:teaser}
\end{figure*}

\section{Introduction}\label{sec:intro}
Video question answering (VideoQA) is a challenging task that requires fine-grained multi-modal understanding.
State-of-the-art approaches to VideoQA~\cite{le2020hierarchical, yu2021learning, zellers2021merlot} rely on large video datasets manually annotated with question-answer pairs.
Yet, collecting such annotations is time consuming, expensive and therefore not scalable.
This has motivated the development of \emph{zero-shot} VideoQA approaches~\cite{yang2021just, Yang2022LearningTA, zellers2022merlot}, 
that use no visual question-answer annotation for training, see Figure~\ref{fig:teaser}.

Recently, a promising line of work builds on \emph{frozen} large autoregressive language models~\cite{eichenberg2021magma,mokady2021clipcap,tsimpoukelli2021multimodal,wang2022language,yang2021empirical,zeng2022socratic} for zero-shot visual question answering.
This has been motivated by the findings from GPT-3~\cite{brown2020language} which exhibits strong zero-shot text-only question answering abilities from large autoregressive language models. 
Such models~\cite{brown2020language, raffel2020exploring, so2021primer, vaswani2017attention} can predict an arbitrarily long sequence of text, one token at each step from left to right.
However, they usually require billion parameters to work well, making them computationally expensive to train, and challenging to deploy in practice.

In contrast, recent work in natural language~\cite{mahabadi2022perfect, schick2020exploiting, schick2020s, tam2021improving} demonstrates strong zero-shot performance for lighter bidirectional language models (BiLM).
Such models~\cite{bert18, he2021deberta, joshi2020spanbert, lan2019albert, liu2019roberta, sanh2019distilbert} can predict a few masked tokens in an input sequence given left and right context in a single forward pass.
These works cast downstream tasks in \emph{cloze} form\footnote{``Cloze test" is an exercise test where certain portions of text are occluded or masked and need to be filled-in.}~\cite{taylor1953cloze}, similar to the masked language modeling task (MLM)~\cite{bert18} solved by these models at pretraining.
This motivates us to tackle diverse zero-shot multi-modal tasks (open-ended VideoQA~\cite{xu2017video}, multiple-choice VideoQA~\cite{lei2018tvqa} and fill-in-the-blank~\cite{maharaj2017dataset}) by formulating them in \emph{cloze} form and leveraging the text-only knowledge of pretrained BiLM.

To adapt a pretrained BiLM to multi-modal inputs, we combine it with a frozen pretrained visual backbone and a set of lightweight additional modules including adapters~\cite{houlsby2019parameter}. 
We train these modules on Web-scraped video-text data using a simple visually-conditioned MLM loss.
We preserve the uni-modal knowledge of a BiLM by \emph{freezing} its weights.
To our knowledge, our approach is the first to explore the zero-shot visual-linguistic capabilities of \emph{frozen non-autoregressive} language models.

We show that our approach largely improves the state of the art on various zero-shot VideoQA benchmarks. 
Furthermore, we demonstrate that \emph{frozen bidirectional} language models perform better while being cheaper to train than \emph{frozen autoregressive} language models~\cite{tsimpoukelli2021multimodal}. 
Moreover, our ablation studies show (i)~the ability of our model to effectively perform zero-shot multi-modal reasoning using both visual cues and speech transcripts, (ii)~the importance of adapters combined with \emph{frozen} pretrained language models, (iii) the impact of multi-modal data scale, (iv)~the impact of the language model size and of bidirectional modeling.
Our approach also performs competitively in the fully-supervised setting. 
Indeed, we show the benefits of \emph{freezing} the weights of a BiLM when using VideoQA training data, while updating considerably less parameters compared to alternative methods. 
Finally, we introduce a new few-shot VideoQA task in which we finetune our pretrained model on a small fraction of the downstream training dataset, and show promising results in this setting.

In summary, our contributions are three-fold:
\vspace{-.3cm}
\begin{itemize}
\item[\textit{(i)}] We present \model{}, a framework that handles multi-modal inputs using \emph{frozen} bidirectional language models and enables zero-shot VideoQA through masked language modeling.\vspace{-.1cm}
\item[\textit{(ii)}] We provide an extensive ablation study and demonstrate the superior performance of our framework in the zero-shot setting when compared to previous autoregressive models.
\vspace{-.1cm}
\item[\textit{(iii)}] Our approach improves the state of the art in zero-shot VideoQA by a significant margin. 
\model{} also demonstrates competitive performance in the fully-supervised setting and shows strong results in the few-shot VideoQA setting which we introduce.
\end{itemize}
\vspace{-.3cm}
Our code and trained models are publicly available at~\cite{frozenbilmwebpage}.

\section{Related Work}\label{sec:background}
\noindent \textbf{Zero-shot VideoQA.} A vast majority of VideoQA approaches rely on relatively small, manually annotated VideoQA datasets~\cite{amrani2020noise, castro2020lifeqa, chadha2020iperceive, choi2020dramaqa, colas2019tutorialvqa, dang2021object, fan2019heterogeneous, gao2018motion, garcia2020knowit, huang2020location, jiang2020divide, jiang2020reasoning, kim2020dense, kim2020modality, kim2017deepstory, kim2021self, le2020hierarchical, le2020neural, lei2019tvqa+, li2019beyond, lin2021vx2text, mun2017marioqa, park2021bridge, sadhu2021video, seo2020look, seo2022end, song2018explore, tapaswi16movieqa, xiao2021next, xue2018better, yang2020bert, ye2017video, zha2019spatiotemporal, zhuang2020multichannel}.
Recently, a few work~\cite{yang2021just, zellers2022merlot} have explored zero-shot approaches for VideoQA, where models are \emph{only} trained on automatically mined video clips with short text descriptions. 
In contrast to VideoQA annotations, such video-text pairs are readily-available at scale on the Web~\cite{bain2021frozen,miech19howto100m, zellers2021merlot}.
In particular, \citet{yang2021just} automatically generate VideoQA training data using language models~\cite{raffel2020exploring} pretrained on a manually annotated text-only question-answer corpus~\cite{rajpurkar2016squad}. 
Reserve~\cite{zellers2022merlot} uses GPT-3~\cite{brown2020language} to rephrase questions into sentences completed by a multi-modal model. 
In contrast to these prior works~\cite{yang2021just,zellers2022merlot}, our method does not require any kind of explicitly annotated language dataset or the use of data generation pipelines for zero-shot VideoQA. 
Note that BLIP~\cite{li2022blip} studies a related setting where a model trained on manually  annotated image-question-answer triplets is transferred to VideoQA, which is a less challenging task.
Also note that VideoCLIP~\cite{xu2021videoclip} considers a related zero-shot multiple-choice video-to-text retrieval task as VideoQA, but in this setting the model is not provided with natural language questions.

\noindent \textbf{Visual language models.} 
As language models require large amounts of training data to perform well~\cite{hoffmann2022training}, recent works have studied transferring pretrained language models~\cite{brown2020language, gpt-j} to image-text tasks.
VisualGPT~\cite{chen2021visualgpt} and VC-GPT~\cite{luo2022vc} showed the benefit of initializing the weights of an image captioning model with a pretrained autoregressive language-only model.
Recent work pushed this idea further by \emph{freezing} the weights of a pretrained autoregressive language model for tackling vision and language tasks~\cite{alayrac2022flamingo, eichenberg2021magma, mokady2021clipcap, tsimpoukelli2021multimodal, wang2022language, yang2021empirical, zeng2022socratic}.
Our approach also leverages a \emph{frozen} pretrained language model.
Similar to MAGMA~\cite{eichenberg2021magma}, we also use adapter layers~\cite{houlsby2019parameter, hu2021lora}.
However, we differ from these approaches as we propose to instead use lighter \emph{bidirectional masked language models}, instead of autoregressive ones, and rely on a masked language modeling objective (MLM) instead of an autoregressive one.
Moreover, our model is specifically designed for videos, for which high-quality visual question answering annotation is even more scarce compared to still images~\cite{eichenberg2021magma, mokady2021clipcap, tsimpoukelli2021multimodal, yang2021empirical}.
We also explore the use of the speech modality, and tackle tasks which are challenging for autoregressive language models such as video-conditioned fill-in-the-blank~\cite{maharaj2017dataset}.
Finally we show in Section~\ref{sec:autoreg} the superior performance of frozen bidirectional language models in comparison with autoregressive ones~\cite{tsimpoukelli2021multimodal}.

\noindent \textbf{Masked Language Modeling in vision and language.} 
The MLM objective was initially introduced in natural language~\cite{bert18, lan2019albert, liu2019roberta} to pretrain bidirectional transformers and learn generic representations.
This approach achieved state-of-the-art results in many language tasks after finetuning on downstream datasets.
Its success inspired numerous works to adapt it to train multi-modal transformer models on paired visual-linguistic data~\cite{chen2019uniter, fu2021violet, gan2020large, hendricks2021decoupling, huang2020pixel, kim2021vilt, lei2021less, li2019unicodervl, li2019visualbert, li2020hero, li2020oscar, li2021align, li2021align2, lu2019vilbert, lu202012, shen2021much, singh2021flava, su2019vl, sun2019videobert, tan2019lxmert, wang2022all, wang2021ufo, yu2020ernie, zellers2021merlot, zhou2020unified, zhu2020actbert}.
However, these works typically use it to learn generic visual-linguistic representations by updating the transformer weights, and then use expensive manual supervision to 
train randomly initialized task-specific answer classifiers for VQA~\cite{chen2019uniter, gan2020large, li2019unicodervl, li2021align, li2020oscar, lu2019vilbert, shen2021much, singh2021flava, su2019vl, tan2019lxmert, wang2021ufo, yu2020ernie} or VideoQA~\cite{fu2021violet, lei2021less, li2021align2, wang2022all, zellers2021merlot}.
In contrast, we tackle \emph{zero-shot} VideoQA, \textit{i.e.}~without using \emph{any} manual annotation. 
Moreover, we do not update the transformer weights during cross-modal training, but instead exhibit the benefits of \emph{freezing} these weights after text-only pretraining, for both zero-shot and fully-supervised VideoQA (see Sections~\ref{sec:ablations} and~\ref{sec:sota}).

\section{Method}\label{sec:method}
This section presents our approach to tackle {\em zero-shot} video question answering. 
Here, zero-shot means that we do not use \emph{any} visual question answering annotation and only rely on scalable data from the Web.
Our approach starts with two strong pretrained components:
(i) a text-only bidirectional masked language model (BiLM) pretrained on data from the Internet, which has the capability of zero-shot question answering but is not capable of visual reasoning, and (ii) a vision encoder pretrained to map images to text descriptions, but which does not have the ability to perform visual question answering.
We aim at connecting these two components while keeping the language component \emph{frozen} to avoid catastrophic forgetting~\cite{DeLange21}, 
where the large language model would specialize to a new task while forgetting its initial capabilities.
The end-goal is to design a unified model having the best of both worlds: visual understanding capabilities of a powerful visual encoder and question answering capabilities of a powerful language model.
This requires several technical innovations, which are described in the rest of this section. 
First, we explain in Section~\ref{sec:architecture} how we augment a \emph{frozen} pretrained bidirectional masked language model with new layers to enable joint video and language reasoning, see Figure~\ref{fig:overview}.
Second, we present in Section~\ref{sec:training} how we train these layers on video-text data scraped from the Web~\citep{bain2021frozen}.
Finally, we describe in Section~\ref{sec:downstream} how we enable zero-shot predictions for several video-language downstream tasks, including open-ended VideoQA, by casting them in a \emph{cloze} form, similar to the masked language modeling task solved during training. 

\subsection{Architecture}\label{sec:architecture}
 
\begin{figure*}[t]
\centering
\includegraphics[width=1.\linewidth]{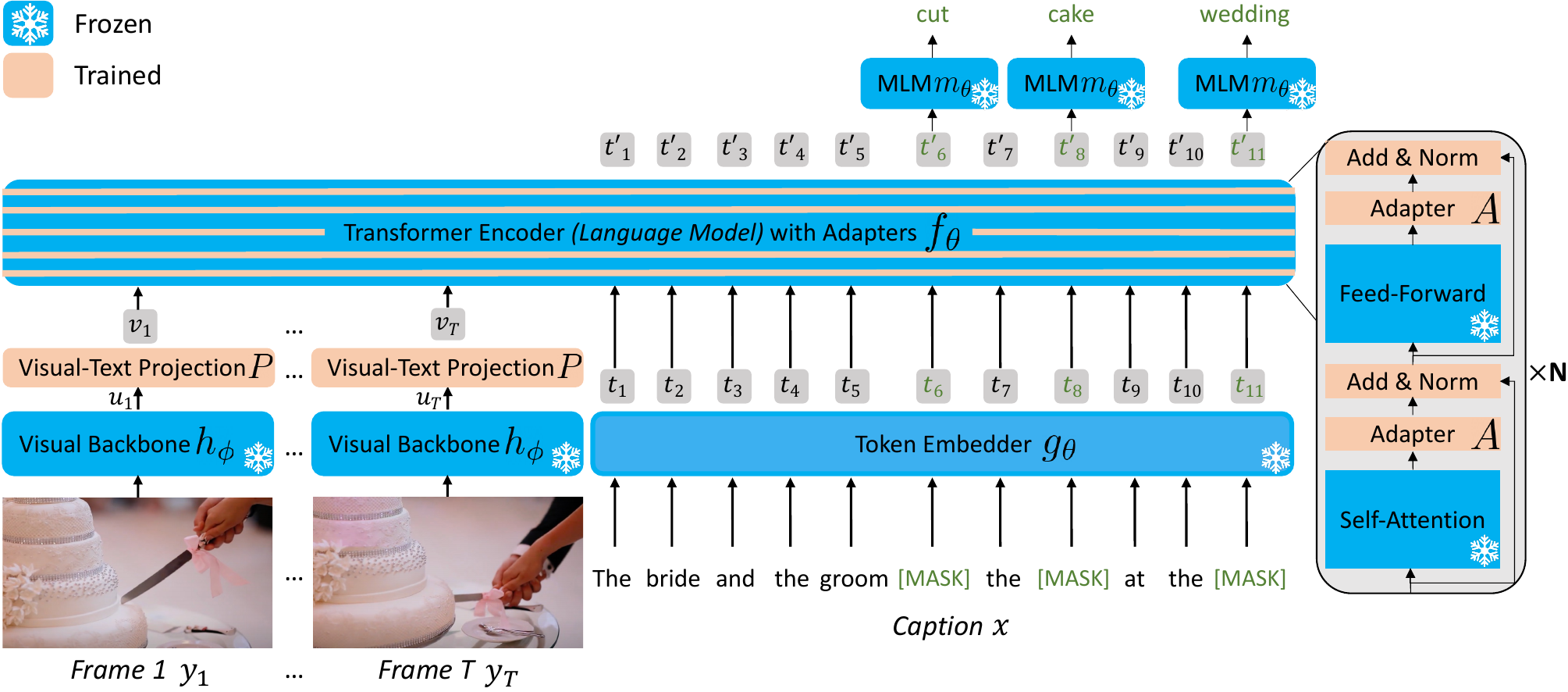}
\caption{\small \textbf{Our training architecture} consists of a large \emph{frozen} bidirectional language model (BiLM) and a \emph{frozen} pretrained visual encoder (in blue), complemented with additional lightweight trainable modules (in orange): (1) a visual-to-text projection module $P$ (on the left), which maps the \emph{frozen} visual features to the joint visual-text embedding space and (2) a set of small adapter modules $A$ (on the right) in between the \emph{frozen} transformer blocks.
The pretrained normalization layers in the BiLM (on the right) are also finetuned. 
}
\label{fig:overview}
\end{figure*}

The proposed architecture, illustrated in Figure~\ref{fig:overview}, brings together a powerful \emph{frozen} pretrained bidirectional language model  with a strong visual encoder. 
The difficulty lies in enabling multi-modal reasoning while keeping the large language model \emph{frozen}.
To address this challenge, we unify these two models via a visual-to-text projection module together with small adapter modules inserted within the frozen language model.
Next, we describe in more detail the three main components of the architecture: (i) the \emph{frozen} pretrained bidirectional language model, (ii) the pretrained video encoder and (iii) the lightweight modules that seamlessly connect the two components. 

\noindent \textbf{\emph{Frozen} Bidirectional Masked Language Model.}
Our method starts from a pretrained  bidirectional language model based on a Transformer encoder~\cite{vaswani2017attention}.
The input text is decomposed into a sequence of tokens $x=\{x_i\}_{1}^{L} \in [1, V]^L$ by a tokenizer of a vocabulary size $V$.
The language model, parameterized by $\theta$, makes use of an embedding function $g_\theta$ which independently transforms each token into a $D$-dimensional continuous embedding $t=\{t_i\}_{1}^{L} := \{g_\theta(x_i)\}_{1}^{L} \in \mathbb{R}^{L\times D}$, a Transformer encoder $f_\theta$ which computes interactions between all input tokens and outputs contextualized representations $t'=\{t'_i\}_{1}^{L}$, and a masked language modeling (MLM) classifier head $m_\theta$ which independently maps the $D$-dimensional continuous embedding for each token $t'_i$ to a vector of logits parameterizing a categorical distribution over the vocabulary $V$. 
This distribution is referred to by ${\log\,p_\theta(x)} := \{m_\theta(t'_i)\}_{1}^{L} \in \mathbb{R}^{L\times V}$.
We assume that the language model is pretrained, \textit{i.e.}~$\theta$ has been optimised with a standard MLM objective~\cite{bert18} on a large dataset of text from the Web.
We show in Section~\ref{sec:ablations} that this text-only pretraining has a crucial importance for zero-shot VideoQA.

\noindent \textbf{Pretrained Video Encoder.} 
The video is represented by a sequence of frames $y=\{y_i\}_{1}^{T}$. 
Each frame is forwarded separately through a visual backbone $h_\phi$, which outputs one feature vector per frame $u = \{u_i\}_{1}^{T} := \{h_\phi(y_i)\}_{1}^{T} \in \mathbb{R}^{T\times D_u}$. 
In detail, the visual backbone is CLIP ViT-L/14~\cite{dosovitskiy2021an, radford2021learning} at resolution $224\times224$ pixels, pretrained to map images to text descriptions with a contrastive loss on 400M Web-scraped image-text pairs. 
The backbone is kept frozen throughout our experiments. 
Note that a CLIP-baseline for zero-shot VideoQA results in poor performance, see Section~\ref{sec:zssota}.

\noindent \textbf{Connecting the Frozen Language and Frozen Vision components.} 
The video features are incorporated into the language model as a prompt~\cite{lester2021power, li2021prefix, zhou2021learning} $v$ of length $T$ (Figure~\ref{fig:overview}, left).
This prompt is obtained by linearly mapping the visual features $u$ to the text token embedding space via a visual-to-text projection $P \in \mathbb{R}^{D_u\times D}$, \textit{i.e.}~$v = \{v_i\}_{1}^{T} := \{P(u_i)\}_{1}^{T}$.
The prompt is then concatenated with the text embeddings before being forwarded to the transformer encoder that models joint visual-linguistic interactions.
We show in Section~\ref{sec:ablations} that incorporating the input video considerably improves zero-shot VideoQA results.
In addition, to learn powerful multi-modal interactions while keeping the transformer encoder weights \emph{frozen}, we equip the transformer encoder with lightweight adapter modules $A$~\cite{houlsby2019parameter} (Figure~\ref{fig:overview}, right). 
We use an adapter which transforms the hidden state $z$ with a multi-layer perceptron transformation and a residual connection, \textit{i.e.}~$A(z) = z + W^{up}\psi(W^{down}z)$ with $W^{down} \in \mathbb{R}^{D\times D_h}$, $W^{up} \in \mathbb{R}^{D_h\times D}$, $D$ the hidden dimension of the transformer, $D_h$ the bottleneck dimension, and $\psi$ a ReLU activation function.
$D_h$ is typically set to be smaller than $D$ such that the adapters are lightweight.
In detail, we add an adapter module before the layer normalization, after each self-attention layer and each feed-forward layer of the transformer encoder. 

\subsection{Cross-modal training}\label{sec:training}

We wish to train the newly added modules introduced in the previous section (shown in orange in Figure~\ref{fig:overview}) for the VideoQA task.
This is hard because we assume that no explicit manual annotation for the VideoQA task is available, such annotations being expensive and therefore hard to obtain at scale.
Instead we train our architecture using \emph{only}  readily-available video-caption pairs scraped from the Web.
Such data is easy to obtain~\cite{bain2021frozen, miech19howto100m, zellers2021merlot}, ensuring the scalability of our approach. 

During training, we keep the weights of the pretrained BiLM and pretrained visual backbone \emph{frozen} as previously explained.
We train from scratch the parameters of (i) the visual-to-text projection module $P$ and (ii) the adapter modules $A$. 
We show in Section~\ref{sec:ablations} the importance of \emph{freezing} the BiLM weights combined with training the adapter modules.
Note that all normalization layers~\cite{ba2016layer} of the pretrained BiLM are also updated to adjust to the new distribution of the training data.
We denote all the trainable parameters of our model by the subscript $\mu$.
In practice, they sum up to about 5\% of the BiLM parameters, hence the training of our model is computationally efficient.

We use a visually-conditioned masked language modeling objective (MLM), in which some text tokens $\{x_{m}\}$ are randomly masked and the model has to predict these tokens based on the surrounding text tokens and the video input. Formally, we minimize the following loss:
\begin{equation}
{\mathcal{L}_\mu(x, y)} = - \frac{1}{M}\sum_{m}{\log\,p_\mu(\tilde{x}, y)_{m}^{x_{m}}},
\end{equation}
where $\tilde{x}$ is the corrupted text sequence, $y$ is the sequence of video frames, $p_\mu(\tilde{x}, y)_{m}^{x_{m}}$ is the probability for the (masked) $m$-th token in $\tilde{x}$ to be $x_{m}$, and $M$ is the number of masks in the sequence $\tilde{x}$.
In detail, we follow~\cite{bert18} and corrupt 15\% of text tokens, replacing them 80\% of the time with a mask token, 10\% of the time with the same token and 10\% of the time with a randomly sampled token.

\subsection{Adapting to downstream tasks}\label{sec:downstream}

After training, our model is able to fill gaps in the input text given an input video together with left and right textual context as part of the input text.
We wish to apply our model \emph{out-of-the-box} to predict an answer given a question about a video.
The video can optionally come with textual subtitles obtained using automatic speech recognition.
To avoid using manual supervision, we formulate the downstream tasks in \emph{cloze} form~\cite{schick2020exploiting, taylor1953cloze}, \textit{i.e.}~such that the model only has to fill-in a mask token in the input prompt similarly to the MLM objective optimized during training. 
The adaptation to the downstream tasks brings several challenges, as described next.
First, we describe how we formulate the input text prompts for several downstream tasks.
Then, we explain how we map the mask token from the input text prompt to an answer via a \emph{frozen} answer embedding module.
Finally, we present how we finetune our architecture in a supervised setting.

\paragraph{Input prompt engineering.} We describe how we design the input text prompts for several downstream video-language tasks. 
Each downstream task is formulated as a masked language modeling problem.
This allows us to apply \model{} out-of-the-box.
A [CLS] token and a [SEP] token are respectively inserted at the start and the end of each sequence following~\cite{bert18}.

\noindent {\em Open-ended VideoQA.} 
Given a question and a video, the task is to find the correct answer in a large vocabulary $\mathcal{A}$ of about 1K answers.
Answers are concise, \textit{i.e.} the great majority of answers consist of one word~\cite{jang2017tgif, xu2017video, yang2021just, yu2019activitynet}.
We design the following prompt: \\
 {\small \texttt{\color{greencode}``[CLS] Question: <Question>? Answer: [MASK]. Subtitles: <Subtitles> [SEP]''}}

\noindent {\em Multiple-choice VideoQA.}
Given a question and a video, the task is to find the correct answer in a small number of candidates $C$, typically up to 5 choices~\cite{lei2018tvqa, li2020hero}. 
We set the vocabulary to $\mathcal{A} = [\textrm{Yes}, \textrm{No}]$ and compute a confidence score for each candidate by using the following prompt: \\
 {\small \texttt{\color{greencode}``[CLS] Question: <Question>? Is it '<Answer Candidate>'? [MASK]. Subtitles: <Subtitles> [SEP]''}}
\\
We choose the best option by selecting the candidate with the highest \textit{Yes} logit value.

\noindent {\em Video-conditioned fill-in-the-blank task.} 
Given a video and a sentence with a blank space, the task is to fill in the blank with the correct word from a vocabulary $\mathcal{A}$ of about 1K answers. 
We replace the blank in the sentence with a mask token, and design the following prompt: \\
 {\small \texttt{\color{greencode}``[CLS] <Sentence with a [MASK] token>. Subtitles: <Subtitles> [SEP]''}}

Note that all prompts are prepended with the video prompt (see Section~\ref{sec:architecture}) before being forwarded to the transformer encoder.

\paragraph{Answer embedding module.}
For each downstream task, we wish to map the mask token in the input text prompt to an actual answer prediction in the set of possible answers $\mathcal{A}$, as described above. 
For this we use the \emph{frozen} MLM classifier head~$m_\theta$.
However, $m_\theta \in \mathbb{R}^{V \times D}$ covers $V$ different tokens where $V>>N$ and $N \approx 1,000$ is the size of $\mathcal{A}$. 
Therefore, we introduce a task-specific answer classification head $l$ which linearly maps a contextualized mask representation $t'_i$ to a vector of logits parameterizing a categorical distribution over the vocabulary $\mathcal{A}$, \textit{i.e.}~$l \in \mathbb{R}^{N \times D}$.
We set the weights of this answer module $l$ with the corresponding weights of the pretrained MLM classifier $m_\theta$ for one-token answers. 
In the case of multi-token answers, we average the weights of their different tokens.
We, hence, enable zero-shot inference at test time.
We also discuss other alternative strategies to handle multi-token answers in Appendix Section~\ref{sec:multitoken}. %

\paragraph{Fully-supervised training.}
To evaluate our approach on fully-supervised benchmarks, we also explore finetuning of our model on datasets that provide manual annotations for the target task.
To this end, we train the same parameters as explained in Section~\ref{sec:training}, %
while keeping
the transformer weights and the answer embedding module \emph{frozen}.
For open-ended VideoQA and video-conditioned fill-in-the-blank, we use a cross-entropy loss on the task-specific vocabulary $\mathcal{A}$.
For multiple-choice VideoQA, we use a binary cross-entropy loss applied to each answer candidate.
We show in Section~\ref{sec:sota} the benefit of \emph{freezing} the language model weights during fully-supervised training.

\section{Experiments}\label{sec:experiments}

This section demonstrates the benefits of our \model{} framework and compares our method to the state of the art.
We first outline our experimental setup in Section~\ref{sec:protocol}.
We then present ablation studies in Section~\ref{sec:ablations}.
Next we compare our bidirectional framework to its autoregressive variant in Section~\ref{sec:autoreg}.
The comparison to the state of the art in zero-shot VideoQA and qualitative results are presented in Section~\ref{sec:zssota}. 
Finally, we finetune our model on the VideoQA task in Section~\ref{sec:sota}, where we show few-shot and fully-supervised results.

\subsection{Experimental setup}\label{sec:protocol}

\noindent \textbf{Frozen bidirectional language model.} 
We use a tokenizer based on SentencePiece~\cite{kudo2018sentencepiece} with $V=128,000$, and a bidirectional language model with 900M parameters, DeBERTa-V2-XLarge~\cite{he2021deberta}, trained with the MLM objective on a corpus of 160G text data.
We also show how our approach generalizes to other MLM-pretrained bidirectional language models such as BERT~\cite{bert18} in Section~\ref{sec:ablations}.

\noindent \textbf{Datasets.} 
For training we use the publicly available \textbf{WebVid10M} dataset~\cite{bain2021frozen}, which consists of 10 million of video-text pairs scraped from the Shutterstock website where video captions are obtained from readily-available alt-text descriptions.
We evaluate results on eight downstream datasets covering a wide range of textual and video domains (\textit{e.g.}~GIFs, YouTube videos, TV shows, movies), and multiple VideoQA paradigms: open-ended VideoQA (\textbf{iVQA}~\cite{yang2021just}, \textbf{MSRVTT-QA}~\cite{xu2017video}, \textbf{MSVD-QA}~\cite{xu2017video}, \textbf{ActivityNet-QA}~\cite{yu2019activitynet} and \textbf{TGIF-QA} FrameQA~\cite{jang2017tgif}), multiple-choice VideoQA (\textbf{How2QA}~\cite{li2020hero} and \textbf{TVQA}~\cite{lei2018tvqa}) and video-conditioned fill-in-the-blank (\textbf{LSMDC}-Fill-in-the-blank~\cite{maharaj2017dataset}).
Unless stated otherwise, we report top-1 test accuracy using the original splits for training, validation and test.
For How2QA, we report results on the public validation set for comparison with prior work~\cite{seo2020look, yang2021just, yu2021learning}.
For TVQA, we report results on the validation set for the ablation studies and on the hidden test set for the comparison to the state of the art.
More details are included in Appendix Section~\ref{sec:adddata}. %

\noindent \textbf{Implementation Details.} 
The training for 2 epochs on WebVid10M lasts 20 hours on 8 Tesla V100 GPUs.
We give further details in Appendix Section~\ref{sec:addimplem}. %

\subsection{Ablation studies}\label{sec:ablations}
In this section, we evaluate the zero-shot performance of different variants of our method.
By default, we use the \emph{frozen} pretrained DeBERTa-V2-XLarge language model and
train the visual-to-text-projection layer together with adapters for 2 epochs on WebVid10M.
We refer to this default model as $\model{}$.
This model uses three input modalities in terms of video, question, and speech.

\noindent \textbf{Ablation of the model training.}
We ablate the effect of initializing parameters of the language model, freezing its weights and training adapters in Table~\ref{table:parameters}.
We observe that the language model pretraining is crucial. 
Indeed, a model with randomly initialized language weights (row 1) performs poorly compared to models initialized with language pretrained weights (rows 2 to 4).
Moreover, the model which updates the language model weights (row 2) during cross-modal training performs considerably worse compared to variants that \emph{freeze} them (rows 3 and 4).
This shows the benefit of \emph{freezing} the language model for zero-shot VideoQA.
We also notice the benefit of the adapter layers by comparing rows 3 and 4, especially for multiple-choice datasets.
Finally, we note that training variants with the \emph{frozen} language model is twice faster compared to updating all parameters, as there is a significantly lower number of parameters to be trained. 

\begin{table}[t]
\begin{center}
\setlength\tabcolsep{3pt}
\resizebox{1.\linewidth}{!}{
\begin{tabular}{lccc|c|ccccc|cc}
& LM & \emph{Frozen} & \multirow{2}{*}{Adapters} &
Fill-in-the-blank &
\multicolumn{5}{c|}{Open-ended} &
\multicolumn{2}{c}{Multiple-choice} \\ 
& Pretraining & LM & &
LSMDC &
iVQA &
MSRVTT-QA & 
MSVD-QA & 
ActivityNet-QA & 
TGIF-QA &
How2QA & 
TVQA
\\ 
\hline
1. & \xmark & \xmark & \xmark & 0.5 & 0.3 & 0.1 & 0.0 & 0.5 & 0.0 & 32.4 & 20.7 \\
2. & \cmark & \xmark & \xmark  & 37.1 & 21.0 & \textbf{17.6} & 31.9 & 20.7 & 30.7 & 45.7 & 45.6 \\ 
3. & \cmark & \cmark & \xmark  & 50.7 & \textbf{27.3} & 16.8 & 32.2 & 24.7 & 41.0 & 53.5 & 53.4 \\
4. & \cmark & \cmark & \cmark  & \textbf{51.5} & 26.8 & 16.7 & \textbf{33.8} & \textbf{25.9} & \textbf{41.9} & \textbf{58.4} & \textbf{59.2} \\
\end{tabular}}
\caption{\small The effect of initializing and training various parts of our model evaluated on zero-shot VideoQA. 
All models are trained on WebVid10M and use multi-modal inputs (video, speech and question) at inference.} 
\label{table:parameters}
\end{center}
\end{table}

\begin{table}[t]
\begin{center}
\setlength\tabcolsep{5pt}
\resizebox{1.\linewidth}{!}{
\begin{tabular}{lcc|c|ccccc|cc}
& \multirow{2}{*}{Visual} & 
\multirow{2}{*}{Speech} & 
Fill-in-the-blank &
\multicolumn{5}{c|}{Open-ended} &
\multicolumn{2}{c}{Multiple-choice} \\ 
& & &
LSMDC &
iVQA &
MSRVTT-QA & 
MSVD-QA & 
ActivityNet-QA & 
TGIF-QA &
How2QA & 
TVQA
\\ 
\hline
1. & \xmark & \xmark & 47.9 & 11.0 & 6.4 & 11.3 & 22.6 & 32.3 & 29.6 & 23.2 \\
2. & \xmark & \cmark & 49.8 & 13.2 & 6.5 & 11.7 & 23.1 & 32.3 & 45.9 & 44.1 \\
3. & \cmark & \xmark & 50.9 & 26.2 & \textbf{16.9} & 33.7 & \textbf{25.9} & \textbf{41.9} & 41.9 & 29.7 \\
4. & \cmark & \cmark & \textbf{51.5} & \textbf{26.8} & 16.7 & \textbf{33.8} & \textbf{25.9} & \textbf{41.9} & \textbf{58.4} & \textbf{59.2} \\
\end{tabular}}
\caption{\small Impact of the visual and speech modalities on zero-shot VideoQA. 
Rows 1 and 2 report results for a pretrained language model without any visual input. 
Rows 3 and 4 give results for a $\textrm{\model{}}$ model pretrained on WebVid10M.}
\label{table:modalities}
\end{center}
\end{table}

\noindent \textbf{Impact of modalities.} 
Table~\ref{table:modalities} shows the impact of the visual and speech modalities on the zero-shot performance of our model.
First, we evaluate the text-only performance of our model using neither visual input nor speech input in row 1. 
We can observe that adding speech (row 2) marginally improves the results and that the importance of speech highly depends on the dataset. 
When adding vision (rows 3 and 4), the performance increases significantly, \textit{e.g.}~+13.6\% accuracy on iVQA and +22.1\% on MSVD-QA between rows 4 and 2.
Finally, the model with vision also benefits from the speech, \textit{e.g.}~+16.5\% accuracy on How2QA and +29.5\% accuracy on TVQA (compare rows 3 and 4).

Note that in practice, speech is missing for many videos, as we obtain the speech directly from the YouTube API and many videos are no longer available. 
Exceptions are How2QA and TVQA for which the authors~\cite{lei2018tvqa, li2021value} provide speech for all videos.
Consequently, we have speech data for only 44.3\%, 14.2\%, 8.2\%, 7.1\% and 25.3\% of test samples in LSMDC-FiB, iVQA, MSRVTT-QA, MSVD-QA and ActivityNet-QA respectively. GIFs in TGIF-QA do not contain speech.

\begin{wraptable}{r}{5.2cm}
\begin{center}
\resizebox{1.\linewidth}{!}{
\begin{tabular}{lc|cc}
& Training Data &
MSVD-QA & 
How2QA
\\ 
\hline
1. & WebVid1K & 13.6 & 53.0 \\
2. & WebVid10K & 22.7 & 54.9 \\
3. & WebVid200K & 27.8 & 56.0 \\
4. & WebVid2M & 30.1 & 57.4 \\ 
5. & WebVid10M & \textbf{33.8} & \textbf{58.4} \\ 
\end{tabular}}
\caption{\small 
Dependency on the size of the training set. Zero-shot results are presented for different fractions of the WebVid10M dataset used for training.
\vspace{-.6cm}
}
\label{table:data}
\end{center}
\end{wraptable}

\begin{table}[t]
\begin{center}
\setlength\tabcolsep{1pt}
\resizebox{1.\linewidth}{!}{
\begin{tabular}{lll|cc|cccccccc}
Method & & Language Model &
\# LM params & \makecell{\small{Train time} \\ \small{(GPUH)}} &
iVQA & 
MSRVTT-QA & 
MSVD-QA & 
ActivityNet-QA & 
TGIF-QA
\\ 
\hline
\multirow{3}{*}{Autoregressive} 
& 1. & GPT-Neo-1.3B & 1.3B & 200 & 6.6 & 4.2 & 10.1 & 17.8 & 14.4 \\
& 2. & GPT-Neo-2.7B & 2.7B & 360 & 9.1 & 7.7 & 17.8 & 17.4 & 20.1 \\
& 3. & GPT-J-6B & 6B & 820 & 21.4 & 9.6 & 26.7 & 24.5 & 37.3 \\
\hline
\multirow{3}{*}{Bidirectional} 
& 4. & BERT-Base & \textbf{110M} & \textbf{24}  & 12.4 & 6.4 & 11.7 & 16.7 & 23.1 \\
& 5. & BERT-Large & 340M & 60  & 12.9 & 7.1 & 13.0 & 19.0 & 21.5 \\
& 6. & DeBERTa-V2-XLarge & 890M & 160 & \textbf{27.3} & \textbf{16.8} & \textbf{32.2} & \textbf{24.7} & \textbf{41.0} \\
\end{tabular}}
\caption{\small Comparison of autoregressive language models (top) and bidirectional language models (bottom) for zero-shot VideoQA.
All variants are trained on WebVid10M for the same number of epochs.}
\label{table:lm}
\end{center}
\end{table}

\noindent \textbf{Size of the cross-modal training dataset.}
Zero-shot results of $\model{}$ after training for a fixed number of iterations on different fractions of WebVid10M are shown in Table~\ref{table:data}.
We construct these subsets such that larger subsets include the smaller ones.
We find that performance increases monotonically with more multi-modal training data.

\noindent \textbf{Size of the language model.} 
In Table~\ref{table:lm}, we ablate the importance of the language model size for the zero-shot performance.
Note that when comparing different language models, we use no adapters to avoid biases related to the choice of the bottleneck dimension hyperparameter~\cite{houlsby2019parameter}.
We find that using the 900M-parameter DeBERTA-V2-XLarge (row 6) outperforms the 300M-parameter BERT-Large (row 5) which also improves over the 100M-parameter BERT-Base (row 4).

\noindent \textbf{Importance of the suffix.}
Our text input prompts include a suffix just to the right of the mask token which consists in a point and an end-of-sentence token for the variant without speech (or a point followed by the speech subtitles for the variant with speech).
We found that removing this suffix leads to a considerable drop of performance (\textit{e.g.}~the test accuracy on MSVD-QA in the row 3 of Table~\ref{table:modalities} drops from 33.7\% to 2.8\%). 
Note that we do not observe such a large drop in performance when removing the [CLS] token \textit{e.g.}~the accuracy on MSVD-QA drops only from 33.8\% to 33.2\%.
This shows that the bidirectional nature of our framework is a key factor for the performance.
Intuitively, this suffix forces the model to provide a concise answer.
Such a hard constraint cannot be given to unidirectional autoregressive models compared next in Section~\ref{sec:autoreg}.
We further ablate the importance of the prompt design in Appendix Section~\ref{sec:addzs}. %

\subsection{Comparison with frozen autoregressive models}\label{sec:autoreg}

In this section, we compare our bidirectional framework using language models of various sizes to the larger, autoregressive GPT-based counterparts recently used for zero-shot image question answering~\cite{tsimpoukelli2021multimodal, yang2021empirical}.
For fair comparison, we adapt autoregressive models to video and language inputs similarly as our bidirectional models.
In detail, autoregressive variants train a similar visual-to-text projection by using a left-to-right language modeling loss~\cite{tsimpoukelli2021multimodal}.
All models in our comparison are trained on WebVid10M for the same number of epochs.
At inference, autoregressive variants use the same template as~\cite{tsimpoukelli2021multimodal} to which we prepend speech subtitles, greedily decode sequences as~\cite{tsimpoukelli2021multimodal}, and use the same answer vocabulary as bidirectional models. 
Autoregressive variants select the top answer that maximizes the log-likelihood when appended to the question prompt.
Here also, we use no adapters for all models, such that the architecture of autoregressive models closely follows~\cite{tsimpoukelli2021multimodal}.
This is to avoid biases related to the tuning of the bottleneck reduction hyperparameter in the adapters~\cite{houlsby2019parameter}. 

We compare autoregressive and bidirectional language models in terms of accuracy and efficiency in Table~\ref{table:lm}.
We observe that our bidirectional framework (rows 4-6) achieves significantly better zero-shot performance-efficiency trade-off compared to its autoregressive counterpart (rows 1-3).
For instance, our framework with BERT-Base~\cite{bert18} (row 4) outperforms the autoregressive variant based on GPT-Neo-1.3B~\cite{gpt-neo} (row 1) which uses 12 times more parameters and 8 times more training time.
Likewise, our framework with DeBERTa-V2-XLarge~\cite{he2021deberta} (row 6) improves over the autoregressive variant based on GPT-J-6B~\cite{gpt-j} (row 3) that has 7 times more parameters and requires 5 times more training time,
showing the efficiency of our \emph{bidirectional} framework for zero-shot VideoQA.

\begin{table}[t]
\begin{center}
\setlength\tabcolsep{1pt}
\resizebox{1.\linewidth}{!}{
\begin{tabular}{lll|c|ccccc|cc}
\multirow{2}{*}{Method} & \multirow{2}{*}{Training Data} &
\multirow{2}{*}{Speech} & Fill-in-the-blank &
\multicolumn{5}{c|}{Open-ended} &
\multicolumn{2}{c}{Multiple-choice} \\ 
& & & LSMDC &
iVQA & 
MSRVTT-QA & 
MSVD-QA & 
ActivityNet-QA & 
TGIF-QA &
How2QA & 
TVQA \\ 
\hline
Random & --- & --- & 0.1 & 0.1 & 0.1 & 0.1 & 0.1 & 0.1 & 25 & 20 \\
CLIP ViT-L/14~\cite{radford2021learning} & 400M image-texts & \xmark & 1.2 & 9.2 & 2.1 & 7.2 & 1.2 & \underline{3.6} & 47.7 & 26.1 \\
Just Ask~\cite{Yang2022LearningTA} & \makecell[l]{HowToVQA69M + \\ WebVidVQA3M} & \xmark & --- & 13.3 & 5.6 & 13.5 & \underline{12.3} & --- & \underline{53.1} & --- \\
Reserve~\cite{zellers2022merlot} & YT-Temporal-1B & \xmark & 31.0 & --- & 5.8 & --- & --- & --- & --- & --- \\
\hline
$\model{}$ (Ours) & WebVid10M & \xmark & \underline{50.9} & \underline{26.2} & \textbf{16.9} & \underline{33.7} & \textbf{25.9} & \textbf{41.9} & 41.9 & \underline{29.7} \\
$\model{}$ (Ours) & WebVid10M & \cmark & \textbf{51.5} & \textbf{26.8} & \underline{16.7} & \textbf{33.8} & \textbf{25.9} & \textbf{41.9} & \textbf{58.4} & \textbf{59.7} \\
\end{tabular}}
\caption{\small Comparison with the state of the art for zero-shot VideoQA.}
\label{table:zeroshot}
\end{center}
\end{table} 

\begin{figure*}[!t]
\centering
\includegraphics[width=1.\linewidth]{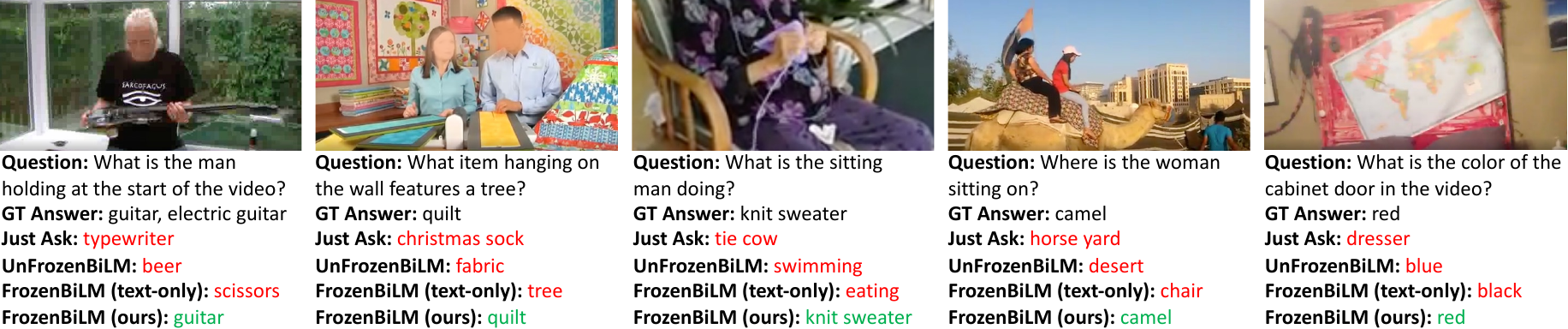}
\caption{\small \textbf{Zero-Shot VideoQA.} Qualitative comparison between Just Ask~\cite{Yang2022LearningTA} (row 3 in Table~\ref{table:zeroshot}), our model (row 4 in Table~\ref{table:zeroshot}), its \textit{unfrozen} variant (row 2 in Table~\ref{table:parameters}) and its text-only variant (row 2 in Table~\ref{table:modalities}).
The first two examples are from iVQA~\cite{yang2021just} and the last three examples are from ActivityNet-QA~\cite{yu2019activitynet}.}
\label{fig:qualitative}
\end{figure*}

\begin{table}[t]
\begin{center}
\setlength\tabcolsep{1pt}
\resizebox{1.\linewidth}{!}{
\begin{tabular}{ll|c|ccccc|cc} %
\multirow{2}{*}{Method} %
& \# Trained & Fill-in-the-blank &
\multicolumn{5}{c|}{Open-ended} &
\multicolumn{2}{c}{Multiple-choice} \\
& Params & LSMDC &
iVQA & 
MSRVTT-QA & 
MSVD-QA & 
ActivityNet-QA & 
TGIF-QA &
How2QA & 
TVQA \\ 
\hline
HCRN~\cite{le2021hierarchical}
& 44M 
& --- & --- & 35.4 & 36.8 & --- & 57.9 
& --- & 71.4\anote \\ 
HERO~\cite{li2020hero}
& 119M
& --- & --- & --- & --- & --- & --- 
& 74.1\anote & 73.6\anote \\
ClipBERT~\cite{lei2021less}
& 114M
& --- & --- & 37.4 & --- & --- & 60.3 
& --- & --- \\ 
Just Ask~\cite{Yang2022LearningTA}
& 157M
& --- & 35.4 & 41.8 & 47.5 & 39.0 & --- 
& 85.3 & --- \\
SiaSamRea~\cite{yu2021learning}
& ---
& --- & --- & 41.6 & 45.5 & 39.8 & 60.2 
& 84.1 & --- \\
MERLOT~\cite{zellers2021merlot}
& 223M
& 52.9 & --- & 43.1 & --- & \underline{41.4} & \textbf{69.5} 
& --- & 78.7\anote \\
Reserve~\cite{zellers2022merlot}
& 644M
& --- & --- & --- & --- & --- & ---
& --- & \textbf{86.1}\anote \\
VIOLET~\cite{fu2021violet}
& 198M
& 53.7 & --- & 43.9 & 47.9 & --- & \underline{68.9} 
& --- & --- \\
All-in-one~\cite{wang2022all}
& 110M
& --- & --- & \underline{46.8} & 48.3 & --- & 66.3 
& --- & --- \\
\hline
\textit{UnFrozenBiLM} (Ours) & 890M & \underline{58.9}\anote & 37.7\anote & 45.0\anote & 53.9\anote & \textbf{43.2}\anote & 66.9 & \textbf{87.5}\anote & 79.6\anote \\
$\model{}$ w/o speech (Ours) & \textbf{30M} & 58.6
& \textbf{39.7} & \textbf{47.0} & \underline{54.4} & \textbf{43.2} & 68.6 & 81.5 & 57.5 \\
$\model{}$ (Ours) & \textbf{30M} & \textbf{63.5}\anote 
& \underline{39.6}\anote & \textbf{47.0}\anote & \textbf{54.8}\anote & \textbf{43.2}\anote & 68.6 & \underline{86.7}\anote & \underline{82.0}\anote \\
\end{tabular}}
\caption{\small Comparison with the state of the art, and the variant \textit{UnFrozenBiLM} which does not freeze the language model weight, on fully-supervised benchmarks.
* denotes results obtained with speech input.}
\label{table:supervised}
\end{center}
\end{table}

\begin{table}[!htbp]
\begin{center}
\setlength\tabcolsep{3pt}
\resizebox{1.\linewidth}{!}{
\begin{tabular}{lc|c|ccccc|cc}
& Supervision &
Fill-in-the-blank &
\multicolumn{5}{c|}{Open-ended} &
\multicolumn{2}{c}{Multiple-choice} \\ 
& &
LSMDC &
iVQA &
MSRVTT-QA & 
MSVD-QA & 
ActivityNet-QA & 
TGIF-QA &
How2QA & 
TVQA
\\ 
\hline
1. & 0\% (zero-shot) & 51.5 & 26.8 & 16.7 & 33.8 & 25.9 & 41.9 & 58.4 & 59.7 \\
2. & 1\% (few-shot) & 56.9 & 31.1 & 36.0 & 46.5 & 33.2 & 55.1 & 71.7 & 72.5 \\
3. & 10\% (few-shot) & 59.9 & 35.3 & 41.7 & 51.0 & 37.4 & 61.2 & 75.8 & 77.6 \\
4. & 100\% (fully-supervised) & \textbf{63.5} 
& \textbf{39.6} & \textbf{47.0} & \textbf{54.8} & \textbf{43.2} & \textbf{68.6} & \textbf{86.7} & \textbf{82.0} \\
\end{tabular}}
\caption{\small Few-shot results, by finetuning $\model{}$ using a small fraction of the downstream training dataset.}
\label{table:fewshot}
\end{center}
\end{table}

\subsection{Comparison to the state of the art for zero-shot VideoQA}\label{sec:zssota}

\noindent \textbf{Quantitative comparison.}
Table~\ref{table:zeroshot} presents results of our method in comparison to the state of the art in \textit{zero-shot} VideoQA settings~\cite{yang2021just}, \textit{i.e.}~when using no manually annotated visual data for training.
Our approach outperforms previous methods by a significant margin on all 8 datasets.
In particular, \model{} outperforms Reserve~\cite{zellers2022merlot}, which is trained on one billion YouTube video clips jointly with vision, language and sound,
Just Ask~\cite{Yang2022LearningTA}, which uses large-scale automatically generated VideoQA data, and a CLIP baseline~\cite{radford2021learning} matching the text concatenating question and answer to the middle frame of the video.
Note that \model{} performs competitively even when using no speech input.
Finally, we note that BLIP~\cite{li2022blip} has a different definition of \textit{zero-shot} where a network finetuned on the image-VQA dataset~\cite{antol2015vqa} is evaluated directly on VideoQA datasets. 
Our Appendix presents results where we outperform BLIP~\cite{li2022blip} in their settings (Section~\ref{sec:blip}) and also includes an analysis of results by question type (Section~\ref{sec:qtype}).
In summary, our evaluation shows the excellent performance of our model in the challenging zero-shot setup.

\noindent \textbf{Qualitative results.}
Figure~\ref{fig:qualitative} illustrates qualitative results of zero-shot VideoQA for our \model{} model and compares them to Just Ask~\cite{Yang2022LearningTA}, as well as to variants of our approach that do not \textit{freeze} the language model (\textit{UnFrozenBiLM}) and use no visual modality (text-only), as evaluated in Section~\ref{sec:ablations}.
We observe that the \textit{unfrozen} variant can predict answers that lack text-only commonsense reasoning, \textit{e.g.}~in the third example, it is unlikely that a sitting man is swimming.
The text-only variant does have strong language understanding, but makes visually-unrelated predictions.
In contrast, consistently with our quantitative results, our model \model{} is able to correctly answer various questions, showing both a strong textual commonsense reasoning and a complex multi-modal understanding. 
We show additional qualitative results in Appendix Section~\ref{sec:addquali}. %

\subsection{Freezing the BiLM is also beneficial in supervised settings}\label{sec:sota}
\noindent \textbf{Fully-supervised VideoQA.} 
We next present an evaluation in a supervised setup where we finetune \model{} on a downstream VideoQA task.
We emphasize that we also keep our pretrained language model weights \emph{frozen} all throughout finetuning. 
As shown in Table~\ref{table:supervised}, our approach improves the state of the art on LSMDC-FiB, iVQA, MSRVTT-QA, MSVD-QA, ActivityNet-QA and How2QA.
In particular, \model{} outperforms strong recent baselines such as All-in-one~\cite{wang2022all} on 2/3 datasets, VIOLET~\cite{fu2021violet} on 3/4 datasets and MERLOT~\cite{zellers2021merlot} on 4/5 datasets. %
Our approach has significantly less trainable parameters compared to the state of the art~\cite{fu2021violet, wang2022all, zellers2021merlot} as we \emph{freeze} the weights of the pretrained language model. 
We ablate this major difference in Table~\ref{table:supervised}, and find that our $\model{}$ with the \emph{frozen} language model performs better and trains twice faster compared to \textit{UnFrozenBiLM} where we update the language model during training.
This shows that \emph{freezing} the language model is not only beneficial for zero-shot but also in fully-supervised settings, therefore suggesting that our \model{} framework also provides a parameter-efficient solution for VideoQA training.
We also note that \model{} performs competitively even without speech input, although speech helps significantly for the performance on LSMDC, How2QA and TVQA.

\noindent \textbf{Few-shot VideoQA.} 
The low number of trainable parameters when training $\model{}$ makes it particularly well-suited in the low data regime.
To verify this, we explore a few-shot VideoQA setting where we finetune our pretrained model using varying fractions of VideoQA training data.
From Table~\ref{table:fewshot} we observe significant improvements over zero-shot when using only 1\% of training data.
Finally, we show in Appendix Section~\ref{sec:addfewshot} that freezing the BiLM highly benefits the few-shot performance, consistently with the results in the zero-shot and fully-supervised settings. %

\section{Conclusion}\label{sec:conclusion}
We have presented \model{}, a framework that extends \emph{frozen} bidirectional language models to multi-modal inputs by training additional modules on Web-scraped data, and that tackles zero-shot VideoQA through masked language modeling.
We have provided extensive ablation studies and shown the efficiency of our framework compared to its autoregressive variant.
\model{} improves the state-of-the-art zero-shot VideoQA on various datasets, performs competitively in fully-supervised settings and exhibits strong performance in the few-shot VideoQA setting we newly introduce.

\noindent \textbf{Limitations.}
Promising directions not explored in this work include scaling the size of a bidirectional language model to several billion parameters, and additional training on large datasets of YouTube videos with accompanying speech transcripts and/or audio~\cite{zellers2022merlot}.
Also, our model cannot be applied out-of-the-box to complex multi-modal text generation tasks such as video captioning.

\noindent \textbf{Broader Impact.}
We have showed the superior compute-efficiency of our bidirectional framework compared to autoregressive models for zero-shot VideoQA, and believe it is a step towards reducing the environmental impact of such research and its applications~\cite{strubell2019energy}.
In addition, our models might reflect biases present in videos and captions from Shutterstock used to train our model, the text data used to train the language model or the images and captions used to train the visual backbone.
It is important to keep this in mind when deploying, analysing and building upon these models.

{\textbf{Acknowledgements.} This work was granted access to the HPC resources of IDRIS under the allocation 2022-AD011011670R2 made by GENCI. The work was funded by a Google gift,  the French government under management of Agence Nationale de la Recherche as part of the "Investissements d'avenir" program, reference ANR-19-P3IA-0001 (PRAIRIE 3IA Institute), the Louis Vuitton ENS Chair on Artificial Intelligence, the European Regional Development Fund under project IMPACT (reg.\ no.\ CZ.02.1.01/0.0/0.0/15 003/0000468).
We thank anonymous reviewers for giving interesting feedback.
We thank Gaspard Beugnot, Clémence Bouvier and Pierre-Louis Guhur for proofreading.}

\newpage

\bibliographystyle{plainnat}  
\bibliography{egbib}

\clearpage \newpage
\appendix

\section*{Appendix}
In this Appendix, we present the following items:
\begin{itemize}
\item[\textit{(i)}] Additional qualitative examples of zero-shot VideoQA predictions (Section~\ref{sec:addquali})
\item[\textit{(ii)}] A qualitative analysis of the \emph{frozen} self-attention patterns in \model{} (Section~\ref{sec:attention})
\item[\textit{(iii)}] Additional information about our experimental setup (Section~\ref{sec:adddetails}), including datasets (Section~\ref{sec:adddata}) and implementation details (Section~\ref{sec:addimplem})
\item[\textit{(iv)}] Additional experimental results (Section~\ref{sec:addexperiments}), including a comparison to BLIP~\cite{li2022blip} in their zero-shot VideoQA settings (Section~\ref{sec:blip}), results on zero-shot image-VQA (Section~\ref{sec:imagevqa}), detailed zero-shot VideoQA results segmented per question type (Section~\ref{sec:qtype}), zero-shot results with different random seeds (Section~\ref{sec:seed}), additional ablation studies in few-shot settings (Section~\ref{sec:addfewshot}), zero-shot settings (Sections~\ref{sec:multitoken} and~\ref{sec:addzs}) and fully-supervised settings (Section~\ref{sec:addablation})
\end{itemize}

\begin{figure*}[!t]
\centering
\begin{subfigure}{1.\linewidth}
\caption{\textbf{Zero-Shot open-ended VideoQA.}
The first row illustrates successful predictions on the iVQA dataset~\cite{yang2021just} (leftmost example) and the ActivityNet-QA dataset~\cite{yu2019activitynet} (three rightmost examples).
The second row illustrates incorrect predictions on the iVQA dataset.}
\label{fig:qualitativesup1}
\vspace{-0.2cm}
\includegraphics[width=1.\linewidth]{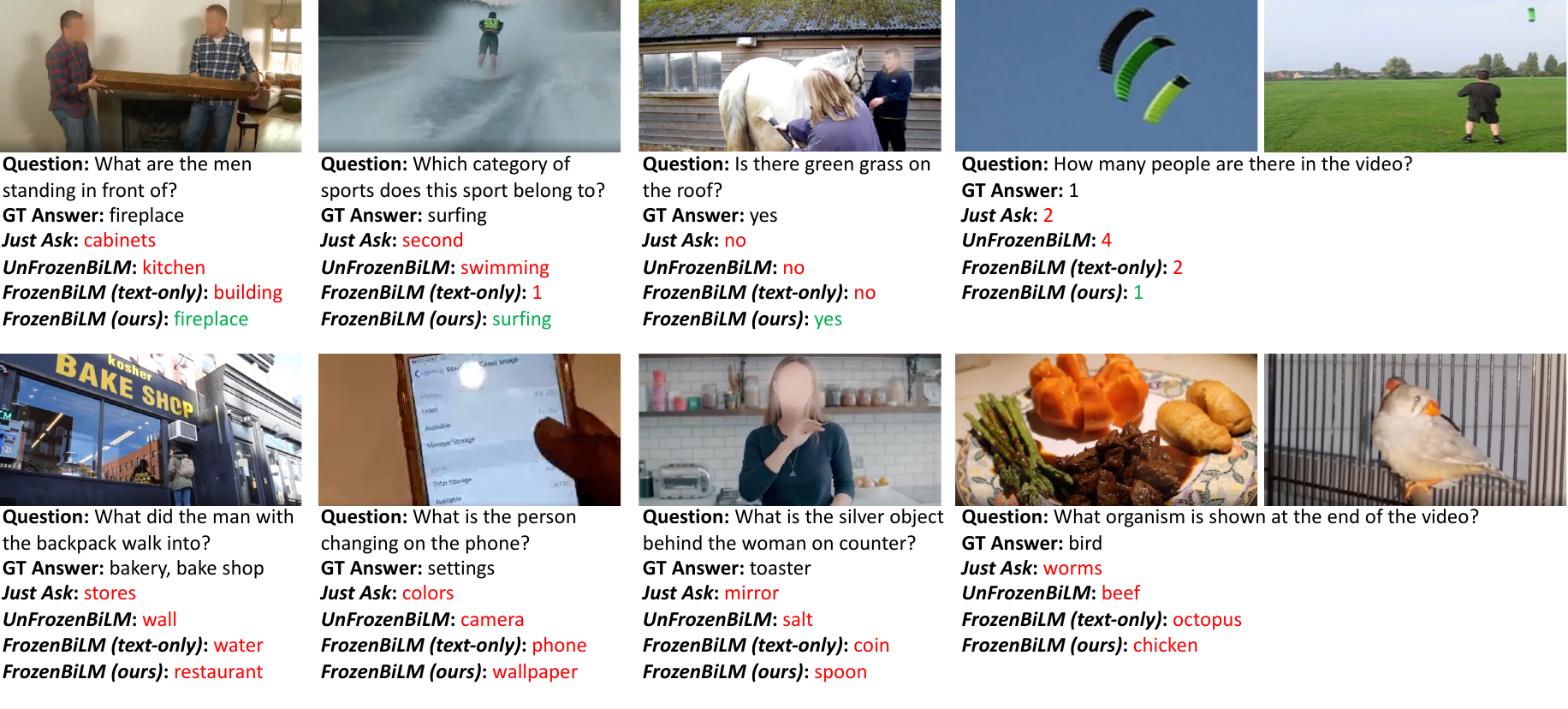}
\end{subfigure}
\begin{subfigure}{1.\linewidth}
\caption{\textbf{Zero-shot video-conditioned fill-in-the-blank} successful predictions on the LSMDC-FiB dataset~\cite{maharaj2017dataset}.}
\label{fig:qualitativesup2}
\vspace{-0.2cm}
\includegraphics[width=1.\linewidth]{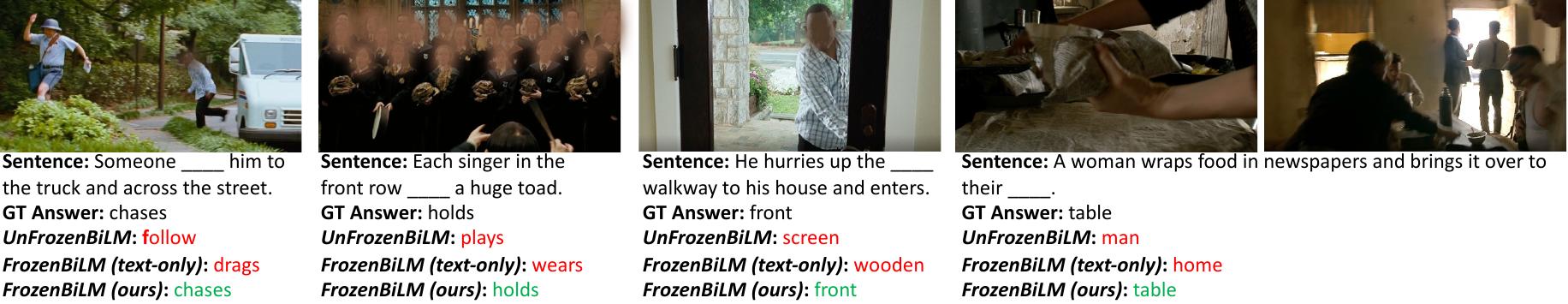}
\end{subfigure}
\begin{subfigure}{1.\linewidth}
\vspace{+0.2cm}
\caption{\textbf{Zero-shot multiple-choice VideoQA.} The first and second rows illustrate successful predictions on the How2QA dataset~\cite{li2020hero} and the TVQA dataset~\cite{lei2018tvqa}, respectively.}
\label{fig:qualitativesup3}
\vspace{-0.2cm}
\includegraphics[width=1.\linewidth]{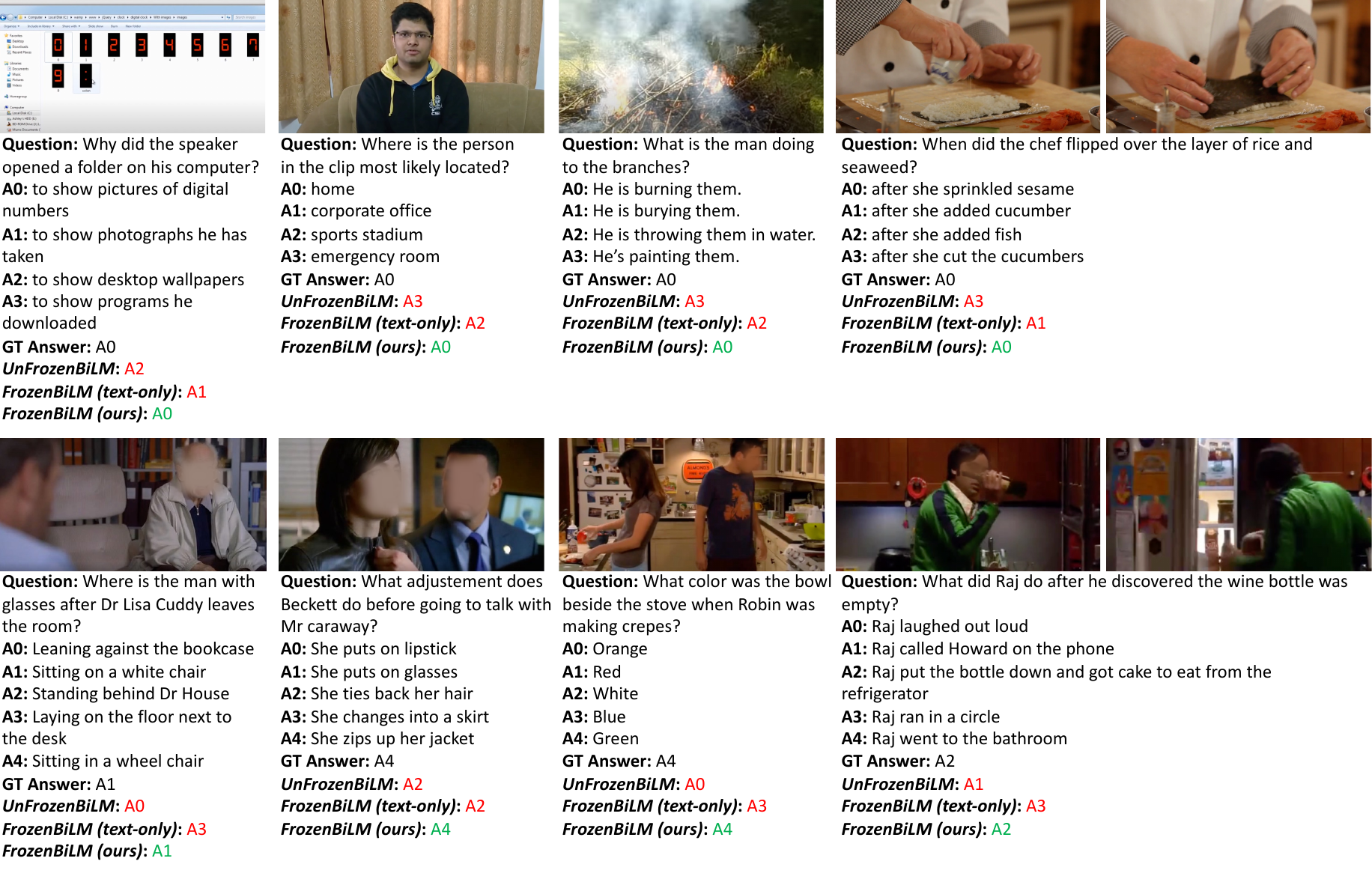}
\end{subfigure}
\vspace{-0.4cm}
\caption{\small \textbf{Zero-Shot VideoQA.} Qualitative comparison between Just Ask~\cite{Yang2022LearningTA} (row 3 in Table~\ref{table:zeroshot}), our model (row 4 in Table~\ref{table:zeroshot}), its \textit{unfrozen} variant (row 2 in Table~\ref{table:parameters}) and its text-only variant (row 2 in Table~\ref{table:modalities}), for zero-shot VideoQA.
The last column of each row illustrates a single video example with two frames, while other columns illustrate each video example with one frame.
We show more examples on our webpage~\cite{frozenbilmwebpage}.}
\label{fig:qualitativesup}
\end{figure*}

\section{Qualitative examples for zero-shot VideoQA}\label{sec:addquali}
To complement the qualitative examples shown in Figure~\ref{fig:qualitative}, Figure~\ref{fig:qualitativesup} and the video \textit{video\_examples.mp4} illustrate additional qualitative results of zero-shot VideoQA for our \model{} model and compares them to Just Ask~\cite{Yang2022LearningTA}, as well as to variants of our approach that do not \textit{freeze} the language model (\textit{UnFrozenBiLM}) and use no visual modality (text-only), as evaluated in Section~\ref{sec:ablations}.
Consistently with the analysis done in Section~\ref{sec:zssota}, we observe that the \textit{unfrozen} variant can predict answers that lack text-only commonsense reasoning, \textit{e.g.}~in the first example of Figure~\ref{fig:qualitativesup2}, the word \textit{follow} is grammatically incorrect; in the second example of Figure~\ref{fig:qualitativesup2}, it is unlikely that a singer \textit{plays} a toad.
The text-only variant does have strong language understanding, but makes visually-unrelated predictions.
In contrast, consistently with our quantitative results (see Tables~\ref{table:parameters}, \ref{table:modalities} and~\ref{table:zeroshot}), our model \model{} is able to correctly answer various questions in the diverse VideoQA paradigms (open-ended VideoQA, video-conditioned fill-in-the-blank, multiple-choice VideoQA), showing both a strong textual commonsense reasoning and a complex multi-modal understanding. 

Our zero-shot model still underperforms compared to VideoQA-supervised models (see Table~\ref{table:fewshot}) and we analyze its failure cases in Figure~\ref{fig:qualitativesup1}.
Qualitatively, we find that the zero-shot model can fail on examples requiring complex temporal or spatial understanding \textit{e.g.}~in the third example of the second row, the model does not detect a toaster behind the woman; in the second example of the second row, it gets confused as the person browses through many different tabs from their phone.
It can also be semantically inaccurate, as in the first example of the second row, the model confuses a restaurant with a bakery; in the fourth example of the second row, it confuses a chicken with another kind of bird.

\begin{figure*}[t]
\centering
\includegraphics[width=1.\linewidth]{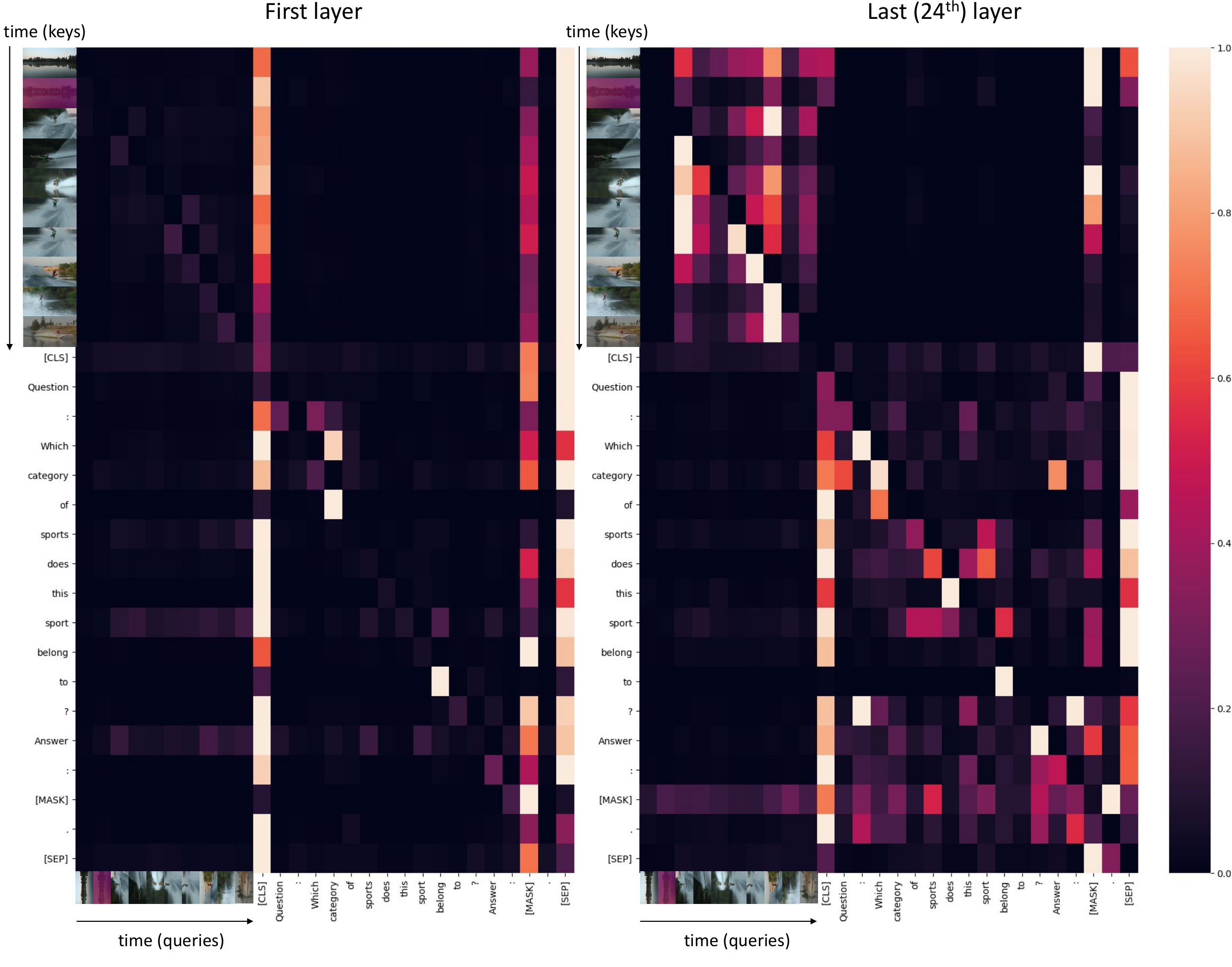}
\caption{\small \textbf{\model{} self-attention visualization for zero-shot VideoQA.} 
Visualization of the attention weights between the different visual tokens from the video prompt and the textual tokens from the text embedder, for the second example of the first row in Figure~\ref{fig:qualitativesup}. 
A column corresponds to the weights of the different visual and text tokens for the given token.
These attention weights are averaged across all 24 heads, and renormalized by the maximum weight for each token (\textit{i.e.}~each column) for the purpose of visualization.
Lighter colors correspond to higher attention
weights (see the colorbar on the right).
In the first layers (left), we observe that the multi-modal interactions mainly flow through the [CLS], [MASK] and [SEP] tokens, and that there is little interaction between the different visual tokens.
In the last layers (right), we observe that visual tokens attend to each other and the [MASK] token attends to the visual tokens, while the [CLS] and [SEP] tokens mainly attend to text tokens.
Note that the self-attention weights are \textit{frozen} after text-only pretraining.}
\label{fig:attnsup}
\end{figure*}

\section{Qualitative analysis of the \emph{frozen} self-attention patterns in \model{}}\label{sec:attention}
We show in Section~\ref{sec:ablations} that the visual modality is crucial for the zero-shot VideoQA performance.
Here we further analyze qualitatively \emph{how}, for zero-shot VideoQA, our model makes use of the visual modality through self-attention layers which are \textit{frozen} after text-only pretraining.
Figure~\ref{fig:attnsup} illustrates the self-attention patterns in \model{} for the second example in the first row of Figure~\ref{fig:qualitativesup}. 
Despite the freezing, we observe that these layers actually enable visual-linguistic interactions.
Indeed, in the first layer (Figure~\ref{fig:qualitativesup}, left), the [CLS], [MASK] and [SEP] tokens significantly attend to the visual tokens.
Moreover, we observe substantially different patterns in the last layer (Figure~\ref{fig:qualitativesup}, right): while the [MASK] token still attends to visual tokens, the different visual tokens at different timesteps attend between each other and the [CLS] and [SEP] tokens mainly attend to other text tokens.
Consistently with results
presented in Section~\ref{sec:ablations}, this qualitative analysis suggests that the \emph{frozen} self-attention layers in \model{} do enable visual-linguistic interactions.

\section{Experimental setup}\label{sec:adddetails}

In this section we first present additional information on the used datasets (Section \ref{sec:adddata}) and then describe implementation details (Section \ref{sec:addimplem}).

\subsection{Datasets}\label{sec:adddata}

In this section, we give further details about the downstream datasets we use.
Their licenses are mentioned in our code in the separate folder \textit{code}.

\noindent \textbf{LSMDC}-FiB~\cite{maharaj2017dataset} is an open-ended video-conditioned fill-in-the-blank task which consists in predicting masked words in sentences that describe short movie clips~\cite{rohrbach15dataset, rohrbach17movie}.
It contains 119K video clips and 349K sentences, split into 297K/22K/30K for training/validation/testing.

\noindent \textbf{iVQA}~\cite{yang2021just} is a recently introduced open-ended VideoQA dataset, focused on objects, scenes and people in instructional videos~\cite{miech19howto100m}.
It excludes non-visual questions, and contains 5 possible correct answers for each question for a detailed evaluation.
It contains 10K video clips and 10K questions, split into 6K/2K/2K for training/validation/testing.

\noindent \textbf{MSRVTT-QA}~\cite{xu2017video}, \textbf{MSVD-QA}~\cite{xu2017video} and \textbf{TGIF-FrameQA}~\cite{jang2017tgif} are popular open-ended VideoQA benchmarks automatically generated from video descriptions~\cite{chen2011collecting, tgif-cvpr2016, xu16msrvtt}.
Questions are of five types for MSRVTT-QA and MSVD-QA: what, who, how, when and where; and four types for TGIF-QA: object, number, color and location.
MSRVTT-QA contains 10K video clips and 243K question-answer pairs, split into 158K/12K/73K for training/validation/testing.
MSVD-QA contains 1.8K video clips and 51K question-answer pairs, split into 32K/6K/13K for training/validation/testing.
TGIF-QA contains 46K GIFs and 53K question-answer pairs, split into 39K/13K for training/testing.

\noindent \textbf{ActivityNet-QA}~\cite{yu2019activitynet} is an open-ended VideoQA dataset consisting of long videos~\cite{caba2015activitynet} (3 minutes long on average), and covering 9 question types (motion, spatial, temporal, yes-no, color, object, location, number and other).
It contains 5.8K videos and 58K question-answer pairs, split into 32K/18K/8K for training/validation/testing.

\noindent \textbf{How2QA}~\cite{li2020hero} is a multiple-choice VideoQA dataset focused on instructional videos~\cite{miech19howto100m}.
Each question is associated with one correct and three incorrect answers. 
It contains 28K video clips and 38K questions, split into 35K/3K for training/validation.

\noindent \textbf{TVQA}~\cite{lei2018tvqa} is a multiple-choice VideoQA dataset focused on popular TV shows.
Each question is associated with one correct and four incorrect answers. 
It contains 22K video clips and 153K questions, split into 122K/15K/15K for training/validation/testing.
The test set is hidden and only accessible a limited number of times via an online leaderboard.

\subsection{Implementation details}\label{sec:addimplem}

\noindent \textbf{Architecture hyperparameters.} 
We truncate text sequences up to $L=256$ tokens.
Video features are extracted by sampling $T=10$ frames, each resized at $224 \times 224$ pixels, from the video.
These frames are sampled at temporally equal distance, with a minimum distance of 1 second.
For videos shorter than $T$ seconds, we pad the video prompt up to $T$ tokens.
The dimension of the visual features from ViT-L/14~\cite{dosovitskiy2021an} is $D_f=768$.
The transformer encoder from DeBERTa-V2-XLarge~\cite{he2021deberta} has 24 layers, 24 attention heads, a hidden dimension of $D=1536$ and an intermediate dimension in the feed-forward layers of 6144.
For the adapters~\cite{houlsby2019parameter}, we use a bottleneck dimension of $D_h=\frac{D}{8}=192$.

\noindent \textbf{Training.} 
For all training experiments, we use the Adam optimizer~\cite{kingma15adam} with $\beta=(0.9, 0.95)$ and no weight decay.
We use Dropout~\cite{srivastava2014dropout} with probability $0.1$ in the adapters and in the transformer encoder.
When finetuning the language model weights, we divide the batch size by a factor 2 so to accommodate with the GPU memory constraints.

\noindent \textbf{Cross-modal training.}
To train on WebVid10M, we use a total batch size of 128 video-caption pairs split in 8 NVIDIA Tesla V100 GPUs.
We use a fixed learning rate of $3e^{-5}$ for the variant with adapters.
We find that the variant without adapters that freezes the language model weights prefers a higher learning rate of $3e^{-4}$, and that the variant \textit{UnfrozenBiLM} that finetunes the language model weights prefers a lower one of $1e^{-5}$.

\noindent \textbf{Downstream task finetuning.}
To finetune our model on downstream datasets, we use a total batch size of 32 video-question-answer triplets (respectively 32 video-sentence pairs) split in 4 NVIDIA Tesla V100 GPUs for open-ended VideoQA datasets (respectively video-conditioned fill-in-the-blank datasets) and 16 video-question pairs split in 8 NVIDIA Tesla V100 GPUs for multiple-choice VideoQA datasets.
We train for 20 epochs for all downstream datasets except LSMDC-FiB for which we find that training for 5 epochs leads to similar validation results.
We warm up the learning rate linearly for the first 10\% of iterations, followed by a linear decay of the learning rate (down to 0) for the remaining 90\%. 
On each dataset, we run a random search and select the learning rate based on the best validation results.
We search over 10 learning rates in the range [$1e^{-5}$, $1e^{-4}$] for variants that freeze the language model weights, and [$5e^{-6}$, $5e^{-5}$] for the variant \textit{UnfrozenBiLM} that finetunes the language model weights.

\noindent \textbf{Answer vocabulary for open-ended VideoQA.}
In the zero-shot setting, we use an answer vocabulary composed of the top $1,000$ answers in the corresponding training dataset, following~\cite{zellers2021merlot}. 
In the fully-supervised setting, we experiment both with the vocabulary composed of the top $1,000$ answers and the vocabulary composed of all answers appearing at least twice in the corresponding training dataset and choose the one leading to best validation results.
Following~\cite{zellers2021merlot}, questions with out-of-vocabulary answer are not used for finetuning, and are automatically considered as incorrect during evaluation.

\section{Experiments}\label{sec:addexperiments}

In this section, we complement the experiments presented in Section~\ref{sec:experiments}.
We first present a comparison with BLIP~\cite{li2022blip} in their zero-shot settings in Section~\ref{sec:blip}.
In Section~\ref{sec:qtype} we show detailed zero-shot VideoQA results segmented per question category and compare our method with Just Ask~\cite{yang2021just}.
Next we analyze the impact of the random seed used in the cross-modal training on the zero-shot VideoQA results in Section~\ref{sec:seed}.
We also show the importance of freezing the language model in few-shot settings in Section~\ref{sec:addfewshot}.
We present additional ablation studies in the zero-shot setting in Section~\ref{sec:addzs}.
Finally we show the benefit of cross-modal training and adapter training in fully-supervised settings in Section~\ref{sec:addablation}.

\begin{table}[t]
\begin{center}
\setlength\tabcolsep{1pt}
\resizebox{1.\linewidth}{!}{
\begin{tabular}{lll|ccccc}
Method & Pretraining Data & Finetuning Data & iVQA & MSRVTT-QA & MSVD-QA & ActivityNet-QA & TGIF-QA \\
\hline
BLIP~\cite{li2022blip} & 129M image-text pairs & VQA & --- & 19.2 & 35.2 & --- & --- \\
\model{} (no image-VQA training) & WebVid10M & $\emptyset$ & 26.8 & 16.7 & 33.8 & 25.9 & 41.9 \\ 
\model{} (no cross-modal training) & $\emptyset$ & VQA & 14.6 & 6.9 & 12.6 & 22.6 & 33.3 \\
\model{} (Ours) & WebVid10M & VQA & \textbf{34.6} & \textbf{22.2} & \textbf{39.0} & \textbf{33.1} & \textbf{43.4} \\
\end{tabular}}
\vspace{+0.2cm}
\caption{\small Results of our model after cross-modal training, finetuning on the open-ended image-VQA dataset~\cite{antol2015vqa} and directly evaluating on open-ended VideoQA without using any VideoQA supervision, as in BLIP~\cite{li2022blip}.}
\label{table:img2vid}
\end{center}
\end{table}

\begin{table*}[t]
\setlength\tabcolsep{10pt}
\begin{center}
\resizebox{1.\linewidth}{!}{	
\begin{tabular}{l|ccccccccc}
Method & Motion & Spatial & Temporal & Yes-No & Color & Object & Location & Number & Other \\ %
\hline
Just Ask~\cite{yang2021just} & 2.3 & 1.1 & 0.3 & 36.3 & 11.3 & 4.1 & 6.5 & 0.2 & 4.7 \\ %
\model{} & \textbf{12.7} & \textbf{6.8} & \textbf{1.6} & \textbf{53.2} & \textbf{16.5} & \textbf{17.9} & \textbf{18.1} & \textbf{26.2} & \textbf{25.8} \\ %
\end{tabular}
}
\caption{\small Zero-shot VideoQA results segmented per question type on the ActivityNet-QA dataset, compared with Just Ask~\cite{yang2021just}.}
\label{table:qtypeact}
\end{center}
\end{table*}

\begin{table*}[!t]
\begin{center}
\resizebox{1.\linewidth}{!}{	
\begin{tabular}{l|cccccc|cccccc}
Method & \multicolumn{6}{c}{MSRVTT-QA} & \multicolumn{6}{c}{MSVD-QA} 
\\ 
& What & Who & Number & Color & When & Where %
& What & Who & Number & Color & When & Where \\ \hline %
Just Ask~\cite{yang2021just} & 1.8 & 0.7 & \textbf{66.3} & 0.6 & 0.6 & 4.5 
& 7.8 & 1.7 & \textbf{74.3} & 18.8 & 3.5 & 0.0 \\ %
\model{} & \textbf{10.7} & \textbf{28.7} & 55.0 & \textbf{11.4} & \textbf{9.2} & \textbf{9.3} %
& \textbf{26.0} & \textbf{45.0} & 69.9 & \textbf{56.3} & \textbf{5.2} & \textbf{17.9} \\ %
\end{tabular}
}
\caption{\small Zero-shot VideoQA results segmented per question type on the MSRVTT-QA dataset (left) and the MSVD-QA dataset (right), compared with Just Ask~\cite{yang2021just}.}
\label{table:qtype}
\end{center}
\end{table*}

\subsection{Comparison with BLIP}\label{sec:blip}
In addition to the zero-shot results presented in Section~\ref{sec:zssota}, we here investigate a different but related \textit{zero-shot} setting defined in BLIP~\cite{li2022blip}, where a network trained on manually annotated image-VQA annotations is evaluated directly on open-ended VideoQA datasets. 
In detail, BLIP uses the open-ended image-VQA dataset~\cite{antol2015vqa} for finetuning after pretraining on 129M image-text pairs, 
including COCO~\cite{chen2015microsoft} and Visual Genome~\cite{visualgenome} which are manually annotated.
To adapt our model to this setting, we finetune our model \model{} pretrained on WebVid10M on the image-VQA dataset using the same procedure as for finetuning on VideoQA datasets (see Section~\ref{sec:downstream}), \textit{i.e.}~notably with a \emph{frozen} language model.
In particular, we finetune on VQA for 10 epochs with an initial learning rate of $1e^{-5}$ which is warmed up for the first 10\% iterations, and linearly decayed to 0 for the remaining 90\% iterations.
Table~\ref{table:img2vid} shows that the resulting model not only improves over our model without image-VQA finetuning (\textit{i.e.}~in zero-shot mode as defined in Section~\ref{sec:intro}) or our model trained on VQA only (\textit{i.e.}~without cross-modal training), but also substantially outperforms BLIP on both MSRVTT-QA and MSVD-QA.
These results further demonstrate the strong capabilities of \model{} in settings where no VideoQA annotation is available.

\subsection{Results on zero-shot image-VQA}\label{sec:imagevqa}
We next evaluate our pretrained model on the VQAv2~\cite{antol2015vqa} validation set in the zero-shot setting, \textit{i.e.}, without any supervision of visual questions and answers. Frozen~\cite{tsimpoukelli2021multimodal} achieves 29.5\% accuracy in this setting using an autoregressive language model. In comparison, our \model{} model is 7 times smaller than Frozen and achieves 45.0\% accuracy. We conclude that our model can perform competitively on the image-VQA tasks despite being tailored for videos.

\subsection{Detailed zero-shot VideoQA results segmented per question category}\label{sec:qtype}
We complement the comparison to the state of the art for zero-shot VideoQA given in Section~\ref{sec:zssota} with results segmented per question type for ActivityNet-QA in Table~\ref{table:qtypeact}, and for MSRVTT-QA and MSVD-QA in Table~\ref{table:qtype}.
Compared to Just Ask~\cite{yang2021just}, we observe large and consistent improvements over all question categories, except for the \textit{number} category on MSRVTT-QA and MSVD-QA.
These results show that our approach is efficient in the diverse question categories of zero-shot VideoQA.

\begin{table}[t]
\begin{center}
\setlength\tabcolsep{1pt}
\resizebox{1.\linewidth}{!}{
\begin{tabular}{ll|c|ccccc|cc}
\multirow{2}{*}{Method} & \multirow{2}{*}{Training Data} & Fill-in-the-blank &
\multicolumn{5}{c|}{Open-ended} &
\multicolumn{2}{c}{Multiple-choice} \\ 
& & LSMDC &
iVQA & 
MSRVTT-QA & 
MSVD-QA & 
ActivityNet-QA & 
TGIF-QA &
How2QA & 
TVQA \\ 
\hline
Random & --- & 0.1 & 0.1 & 0.1 & 0.1 & 0.1 & 0.1 & 25 & 20 \\
CLIP ViT-L/14~\cite{radford2021learning} & 400M image-texts & 1.2 & 9.2 & 2.1 & 7.2 & 1.2 & \underline{3.6} & 47.7 & \underline{26.3} \\
Just Ask~\cite{Yang2022LearningTA} & \makecell[l]{HowToVQA69M + \\ WebVidVQA3M} & --- & \underline{13.3} & 5.6 & \underline{13.5} & \underline{12.3} & --- & \underline{53.1} & --- \\
Reserve~\cite{zellers2022merlot} & YT-Temporal-1B & \underline{31.0} & --- & \underline{5.8} & --- & --- & --- & --- & --- \\
$\model{}$ (Ours) & WebVid10M & \textbf{51.5$\pm$0.1} & \textbf{28.3$\pm$0.9} & \textbf{14.4$\pm$1.4} & \textbf{30.0$\pm$2.2} & \textbf{25.4$\pm$0.7} & \textbf{39.7$\pm$2.1} & \textbf{57.9$\pm$0.6} & \textbf{57.9$\pm$1.2} \\ %
\end{tabular}}
\vspace{+0.2cm}
\caption{\small Comparison with the state of the art for zero-shot VideoQA, reporting mean and standard deviation over 5 cross-modal training runs with different random seeds.
Results on TVQA are reported on the validation set given that the hidden test set can only be accessed a limited number of times.}
\label{table:variance}
\end{center}
\end{table} 

\begin{table}[!t]
\begin{center}
\setlength\tabcolsep{1pt}
\resizebox{1.\linewidth}{!}{
\begin{tabular}{lcc|c|ccccc|cc}
& Variant & Supervision &
Fill-in-the-blank &
\multicolumn{5}{c|}{Open-ended} &
\multicolumn{2}{c}{Multiple-choice} \\ 
& & &
LSMDC &
iVQA &
MSRVTT-QA & 
MSVD-QA & 
ActivityNet-QA & 
TGIF-QA &
How2QA & 
TVQA
\\ 
\hline
1. & \textit{UnFrozenBiLM} & 0\% (zero-shot) & 37.1 & 21.0 & \textbf{17.6} & 31.9 & 20.7 & 30.7 & 45.7 & 45.6 \\
2. & \model{} & 0\% (zero-shot) & \textbf{51.5} & \textbf{26.8} & 16.7 & \textbf{33.8} & \textbf{25.9} & \textbf{41.9} & \textbf{58.4} & \textbf{59.2} \\
\hline
3. & \textit{UnFrozenBiLM} & 1\% (few-shot) & 46.2 & 23.5 & 33.4 & 43.7 & 31.6 & 51.7 & 68.0 & 68.6 \\ %
4. & \model{} & 1\% (few-shot) & \textbf{56.9} & \textbf{31.1} & \textbf{36.0} & \textbf{46.5} & \textbf{33.2} & \textbf{55.1} & \textbf{71.7} & \textbf{71.8} \\ %
\hline
5. & \textit{UnFrozenBiLM} & 10\% (few-shot) & 52.6 & 29.5 & 38.9 & 49.8 & 36.5 & 57.8 & 73.2 & 74.8 \\ %
6. & \model{} & 10\% (few-shot) & \textbf{59.9} & \textbf{35.3} & \textbf{41.7} & \textbf{51.0} & \textbf{37.4} & \textbf{61.2} & \textbf{75.8} & \textbf{77.3} \\ %
\hline
7. & \textit{UnFrozenBiLM} & 100\% (fully-supervised) & 58.9
& 37.7 & 45.0 & 53.9 & \textbf{43.2} & 66.9 & \textbf{87.5} & 79.1 \\ 
8. & \model{} & 100\% (fully-supervised) & \textbf{63.5} 
& \textbf{39.6} & \textbf{47.0} & \textbf{54.8} & \textbf{43.2} & \textbf{68.6} & 86.7 & \textbf{82.4} \\
\end{tabular}}
\vspace{+0.2cm}
\caption{\small Few-shot results, by finetuning $\model{}$ using a small fraction of the downstream training dataset, compared with the variant \textit{UnFrozenBiLM} which does not freeze the language model weights.
Results on TVQA are reported on the validation set given that the hidden test set can only be accessed a limited number of times.}
\label{table:addfewshot}
\end{center}
\end{table}

\subsection{Impact of the random seed on zero-shot VideoQA}\label{sec:seed}
To verify the robustness of our approach with respect to the random seed, we run cross-modal training for \model{} with 5 different random seeds.
We report the mean and standard deviation of zero-shot accuracy in Table~\ref{table:variance}, compared with state-of-the-art approaches that only report their results based on a single run.
We observe that the random seed does not affect the comparison to prior work done in Section~\ref{sec:zssota} in the main paper, as our model improves over previous work for zero-shot VideoQA~\cite{radford2021learning, Yang2022LearningTA, zellers2022merlot} by significant margins.

\subsection{Freezing the language model is also beneficial in few-shot settings}\label{sec:addfewshot}
Sections~\ref{sec:ablations} and~\ref{sec:sota} demonstrate that freezing the language model combined with training adapters outperforms finetuning the language model in the zero-shot and fully-supervised settings.
In Table~\ref{table:addfewshot}, we further show that freezing the language model combined with training adapters outperforms finetuning the language model in the few-shot setting as defined in Section~\ref{sec:sota} (compare rows 3 and 4, or rows 5 and 6).
Interestingly, the difference is larger when using 1\% of the downstream training dataset (rows 3 and 4) compared to using 10\% (rows 5 and 6) or 100\% (rows 7 and 8).
These results demonstrate that our approach is particularly efficient in settings where VideoQA annotations are scarce. 

\begin{table}[t]
\begin{center}
\setlength\tabcolsep{6pt}
\resizebox{1.\linewidth}{!}{
\begin{tabular}{ll|c|ccccc}
& Inference Strategy &
Fill-in-the-blank &
\multicolumn{5}{c}{Open-ended} \\ 
& & LSMDC & iVQA & MSRVTT-QA & MSVD-QA & ActivityNet-QA & TGIF-QA \\
\hline
1. & Average token embeddings & \textbf{51.5} & 26.8 & 16.7 & 33.8 & 25.9 & 41.9 \\ 
2. & Multiple mask tokens & 51.0 & \textbf{27.0} & \textbf{17.1} & \textbf{34.4} & \textbf{26.1} & \textbf{42.0} \\
\end{tabular}}
\vspace{+0.2cm}
\caption{\small Impact of the inference strategy on the zero-shot open-ended VideoQA performance.}
\label{table:multitoken}
\end{center}
\end{table}

\subsection{Ablation of the multi-token inference strategy}\label{sec:multitoken}
For multi-token answers in the open ended VideoQA setting, our \model{} simply averages the weights of different answer tokens.
However, such simple scheme does not preserve the semantic structure of the answer.
Hence we here investigate and compare another possible inference strategy in the zero-shot setting and discuss potential sources of improvement.
We take inspiration from \cite{jiang2020x} and performs zero-shot VideoQA inference by using multiple mask tokens decoded in parallel. 
Then, for each video-question pair, we do one forward pass through the model per possible number of mask tokens (typically, 1 to 5) in order to score all possible answers in vocabulary $\mathcal{A}$. 
The score of a given answer is then obtained by multiplying the probability of its individual tokens, possibly normalized by its number of tokens. 
As shown in Table~\ref{table:multitoken}, we observe that such a decoding strategy (row 2) does not significantly improve the accuracy of our model over the one used in \model{} (row 1). 
We hypothesize that this is due to the fact that the current open-ended VideoQA datasets~\cite{jang2017tgif, xu2017video, yang2021just, yu2019activitynet} contain a great majority of short answers, e.g. 99\% of the answers in the MSRVTT-QA test set are one-token long with our tokenizer~\cite{kudo2018sentencepiece}. 
Additionally, a possible solution to further improve the decoding in this alternative scheme is to increase the length of the masked spans at pretraining, as in \cite{joshi2020spanbert}. 
\cite{salazar2019masked} provides another potential solution to score multi-token answers in our framework, by masking tokens one by one and computing pseudo-likelihood scores.

\begin{table}[t]
\begin{center}
\setlength\tabcolsep{4pt}
\resizebox{1.\linewidth}{!}{
\begin{tabular}{lccc|c|ccccc|cc}
& T & $D_h$ & Visual &
Fill-in-the-blank &
\multicolumn{5}{c|}{Open-ended} &
\multicolumn{2}{c}{Multiple-choice} \\ 
& & & Backbone &
LSMDC &
iVQA &
MSRVTT-QA & 
MSVD-QA & 
ActivityNet-QA & 
TGIF-QA &
How2QA & 
TVQA
\\ 
\hline
1. & 1 & 192 & ViT-L/14 (CLIP) & 50.4 & 24.8 & 12.4 & 28.3 & 24.9 & 41.5 & 54.3 & 54.6 \\
2. & 10 & 96 & ViT-L/14 (CLIP) & \textbf{52.4} & \textbf{28.6} & 13.7 & 29.0 & 25.1 & \textbf{42.3} & \textbf{59.3} & 58.0 \\
3. & 10 & 384 & ViT-L/14 (CLIP) & 51.4 & 27.5 & 15.6 & 31.2 & 23.9 & 41.8 & 58.0 & 57.8 \\
4. & 10 & 192 & ViT-B/16 (ImageNet) & 49.4 & 23.8 & 13.3 & 25.7 & 25.1 & 36.8 & 56.5 & 57.2 \\
5. & 10 & 192 & ViT-B/16 (CLIP) & 50.8 & 25.5 & 14.6 & 30.3 & 25.6 & 41.0 & 57.6 & 58.2 \\
6. & 10 & 192 & ViT-L/14 (CLIP) & 51.5 & 26.8 & \textbf{16.7} & \textbf{33.8} & \textbf{25.9} & 41.9 & 58.4 & \textbf{59.2} \\
\end{tabular}}
\vspace{+0.2cm}
\caption{\small Impact of the number of frames $T$ used by the model, the hidden dimension $D_h$ in the adapters and the visual backbone on the zero-shot VideoQA results.
All models are trained on WebVid10M and use multi-modal inputs (video, speech and question) at inference.} 
\label{table:addzsablation}
\end{center}
\end{table}

\begin{table}[t]
\begin{center}
\setlength\tabcolsep{1pt}
\resizebox{1.\linewidth}{!}{
\begin{tabular}{ll|ccccc}
& Template & iVQA & MSRVTT-QA & MSVD-QA & ActivityNet-QA & TGIF-QA \\
\hline
1. & \texttt{\color{greencode}``[CLS] Question: <Question>? Answer: [MASK]. Subtitles: <Subtitles> [SEP]``} & 26.8 & \textbf{16.7} & \textbf{33.8} & \textbf{25.9} & \textbf{41.9} \\ 
2. & \texttt{\color{greencode}``[CLS] Q: <Question>? A: [MASK]. S: <Subtitles> [SEP]``} & \textbf{27.4} & 16.2 & 32.5 & 25.5 & \textbf{41.9} \\
3. & \texttt{\color{greencode}``[CLS] <Question>? [MASK]. <Subtitles> [SEP]``} & 23.1 & 13.6 & 28.0 & 21.6 & 25.2 \\
\end{tabular}}
\vspace{+0.2cm}
\caption{\small Impact of the prompt on the zero-shot open-ended VideoQA performance.}
\label{table:promptoe}
\end{center}
\end{table}

\begin{table}[t]
\begin{center}
\setlength\tabcolsep{1pt}
\resizebox{1.\linewidth}{!}{
\begin{tabular}{ll|cc}
& Template & How2QA & TVQA \\
\hline
1. & \texttt{\color{greencode}``[CLS] Question: <Question>? Is it ’’<Answer Candidate>”? [MASK]. Subtitles: <Subtitles> [SEP]``} & \textbf{58.4} & \textbf{59.7} \\ 
2. & \texttt{\color{greencode}``[CLS] Q: <Question>? Is it ’’<Answer Candidate>”? [MASK]. S: <Subtitles> [SEP]``} & 57.7 & 58.2 \\
3. & \texttt{\color{greencode}``[CLS] <Question>? <Answer Candidate>? [MASK]. <Subtitles> [SEP]``} & 47.6 & 55.0 \\
\end{tabular}}
\vspace{+0.2cm}
\caption{\small Impact of the prompt on the zero-shot multiple-choice VideoQA performance.}
\label{table:promptmc}
\end{center}
\end{table}

\subsection{Additional ablation studies in the zero-shot setting}\label{sec:addzs}
We here complement zero-shot ablation studies reported in Section~\ref{sec:ablations}. 
We analyze the impact of the number of frames $T$ used by the model, the hidden dimension in the adapters $D_h$ and the size and pretraining of the visual backbone in Table~\ref{table:addzsablation}.
All models use the same setting as described in Section~\ref{sec:ablations} and detailed in Section~\ref{sec:adddetails}.
We first observe that using 10 frames significantly improves over using a single frame (compare rows 1 and 5).
Next we note that using a hidden dimension of $96$ or $384$ in the adapters instead of $192$ does not change the results significantly (see rows 2, 3 and 6).
Moreover, we find that scaling up the size of the visual backbone is beneficial, as using ViT-L/14 instead of ViT-B/16, both being trained on CLIP~\cite{radford2021learning}, slightly improves the results (compare rows 4 and 6).
Furthermore, we observe that the pretraining of the visual backbone is crucial, as using ViT-B/16 pretrained on 400M image-text pairs from CLIP significantly improves over using ViT-B/16 pretrained on ImageNet-21K, \textit{i.e.}~22M image-label pairs (compare rows 4 and 5).

Finally, we ablate the importance of the prompt design on the zero-shot VideoQA performance.
We report results with alternative prompts in Tables~\ref{table:promptoe} and~\ref{table:promptmc}.
We find that replacing the words “Question”, “Answer” and “Subtitles” by “Q”, “A” and “S”, respectively, in the templates described in Section~\ref{sec:downstream} does not impact the zero-shot VideoQA accuracy (compare rows 2 and 1 in Tables~\ref{table:promptoe} and~\ref{table:promptmc}).
However, completely removing “Question”, “Answer”, “Subtitles” and “is it” in the templates results in a significant drop of performance (compare rows 3 and 1 in Tables~\ref{table:promptoe} and~\ref{table:promptmc}). 
We conclude that it is important to have tokens that link the different textual inputs.

\begin{table}[!t]
\begin{center}
\setlength\tabcolsep{1pt}
\resizebox{1.\linewidth}{!}{
\begin{tabular}{lcccl|c|ccccc|cc} %
& Cross-modal & 
Frozen 
& \multirow{2}{*}{Adapters}
& \# Trained & Fill-in-the-blank &
\multicolumn{5}{c|}{Open-ended} &
\multicolumn{2}{c}{Multiple-choice} \\ %
& Training & LM & & Params & LSMDC &
iVQA & 
MSRVTT-QA & 
MSVD-QA & 
ActivityNet-QA & 
TGIF-QA &
How2QA & 
TVQA \\ 
\hline
1. & \cmark & \xmark & \xmark & 890M & 58.9 & 37.7 & 45.0 & 53.9 & \textbf{43.2} & 66.9 & \textbf{87.5} & 79.1 \\ %
2. & \cmark & \cmark & \xmark & \textbf{1M} & 60.4
& 38.2 & 43.2 & 51.7 & 38.3 & 66.5 & 79.3 & 78.2 \\ %
3. & \xmark & \cmark & \cmark & 30M & 57.1
& 34.3 & 46.2 & 51.9 & 41.8 & 67.4 & 75.8 & 70.8 \\ %
4. & \cmark & \cmark & \cmark & 30M & \textbf{63.5} 
& \textbf{39.6} & \textbf{47.0} & \textbf{54.8} & \textbf{43.2} & \textbf{68.6} & 86.7 & \textbf{82.4} \\
\end{tabular}} %
\vspace{+0.2cm}
\caption{\small Importance of cross-modal training and training various parameters for fully-supervised VideoQA.
All models are finetuned on downstream VideoQA datasets, and use multi-modal inputs (video, speech and question) at inference.}
\label{table:addsupervised}
\end{center}
\end{table}

\subsection{Cross-modal training and adapters are crucial for fully-supervised performance}\label{sec:addablation}
We have examined the impact of cross-modal training and training various parameters of our architecture on the zero-shot VideoQA performance in Section~\ref{sec:ablations}.
In Table~\ref{table:addsupervised}, we complement these ablation studies by analyzing the importance of cross-modal training and training various parameters for the fully-supervised VideoQA performance.
For this, we train on downstream datasets a variant with no adapters, and a variant without cross-modal training, following the same procedure as explained in Section~\ref{sec:downstream} and detailed in Section~\ref{sec:adddetails}.
We find that cross-modal training is significantly beneficial for the fully-supervised setting (compare rows 3 and 4).
Similar to conclusions made in Section~\ref{sec:sota}, training adapters while freezing the language model outperforms finetuning the language model in fully-supervised settings (see rows 1 and 4).
Finally, we note that training adapters has a considerable importance on the performance in fully-supervised settings (compare rows 2 and 4).
These results further demonstrate the strength of our approach in the fully-supervised setup.

\end{document}